\documentclass[review]{elsarticle}
\usepackage{lineno,hyperref}
\hypersetup{hypertex=true,
	colorlinks=true,
	linkcolor=blue,
	anchorcolor=blue,
	citecolor=blue
}
\usepackage{tabularx,booktabs}
\usepackage{multirow}
\modulolinenumbers[5]
\usepackage{graphicx} 
\usepackage{float}
\usepackage{setspace}
\usepackage{multicol}
\usepackage{geometry}
\usepackage{fontawesome}
\usepackage{multirow}
\usepackage{makecell}
\usepackage{array}
\usepackage{bm} 
\usepackage{enumitem}
\usepackage{amsfonts}
\usepackage{amsmath}
\usepackage{subfigure}
\usepackage{mathtools} 
\journal{Journal of \LaTeX\ Templates}
\usepackage{caption}

\begin{document}
\captionsetup[figure]{labelfont={bf},labelformat={default},labelsep=period,name={Fig.}}
\captionsetup[table]{labelfont={bf},labelsep=newline,singlelinecheck=false}
\begin{frontmatter}

\title{Bayesian Physics-Informed Extreme Learning Machine for Forward and Inverse PDE Problems with Noisy Data}

\author[mymainaddress]{Xu Liu}
\ead{liuxu18054448691@126.com}

\author[mymainaddress]{Wen Yao\corref{mycorrespondingauthor}}

\cortext[mycorrespondingauthor]{Corresponding author}
\ead{Wendy0782@126.com}
\author[mymainaddress]{Wei Peng\corref{mycorrespondingauthor}}
\ead{weipeng0098@126.com}

\author[mymainaddress]{Weien Zhou}

\address[mymainaddress]{Defense Innovation Institute, Chinese Academy of Military Science, Beijing 100071, China}

\begin{abstract}
Physics-informed extreme learning machine (PIELM) has recently received significant attention as a rapid version of physics-informed neural network (PINN) for solving partial differential equations (PDEs). The key characteristic is to fix the input layer weights with random values and use Moore–Penrose generalized inverse for the output layer weights. The framework is effective, but it easily suffers from overfitting noisy data and lacks uncertainty quantification for the solution under noise scenarios.
To this end, we develop Bayesian physics-informed extreme learning machine (BPIELM) to solve both forward and inverse linear PDE problems with noisy data in a unified framework. In our framework, a prior probability distribution is introduced in the output layer for extreme learning machine with physic laws and the Bayesian method is used to estimate the posterior of parameters. Besides, for inverse PDE problems, problem parameters considered as new output layer weights are unified in a framework with forward PDE problems. Finally, we demonstrate BPIELM considering both forward problems, including Poisson, 
advection, and diffusion equations, as well as inverse problems, where unknown problem parameters are estimated. The results show that, compared with PIELM, BPIELM quantifies uncertainty arising from noisy data and provides more accurate predictions. In addition, BPIELM is considerably cheaper than PINN in terms of the computational cost.
\end{abstract}

\begin{keyword}
Partial differential equation \sep Uncertainty quantification \sep Extreme learning machine \sep Physics-informed neural network  
\end{keyword}

\end{frontmatter}


\section{Introduction}

How to leverage the existing data to accurately model and predict PDE-based systems remains an important issue, especially for solving ill-posed problems, e.g., with unknown parameters and boundary conditions, for alleviating the influence by noisy data collected from different sensors, and for reducing the computational cost \cite{psaros2022uncertainty, alber2019integrating}. Recently, physics-informed machine learning, as a powerful tool, has attracted increasing attention to addressing those challenges \cite{karniadakis2021physics}. In particular, physics-informed neural network (PINN), a general framework for solving both forward and inverse PDE problems, encodes the underlying physics laws into loss function to constrain the space of admissible solutions and makes predictions for unknown terms given limited data \cite{raissi2017physics, raissi2019physics, basir2021physics}. PINN has been successfully used for solving PDEs or complex PDE-based physics problems in various domains, such as materialogy \cite{he2021physics}, medical diagnosis \cite{kissas2020machine} and hidden fluid mechanics \cite{jin2021nsfnets}, etc.

Quantifying prediction uncertainty associated with, such as noisy data and unknown terms, is important for neural network (NN) based inference \cite{kabir2018neural, meng2021multi}. However, PINN without the built-in uncertainty quantification (UQ) may restrict physical and biological applications that require high reliability. For some UQ problems, generating distributions of output predictions is more important than point estimates of the solution \cite{zhu2019physics}. In this regard, Yang et al. \cite{yang2019adversarial, yang2020physics} quantify and propagate uncertainty in systems by combining generative adversarial network (GAN) and physics-based loss function. The latent variable models are constructed as probabilistic representations and the adversarial inference procedure is utilized to train models. While physics laws of systems are encoded into the loss function instead of the discriminator. To exploit the full potential of the adversarial inference procedure, Daw et al. \cite{daw2021pid} propose a GAN framework based on a physics-informed discriminator, where the physics laws are incorporated into the learning of both the generator and the discriminator model. In addition, the Bayesian method achieves many successful research efforts about UQ in deep learning \cite{jospin2020hands, goan2020bayesian, wilson2020bayesian}. Yang et al. \cite{yang2021b} combine the Bayesian neural network and PINN to provide predictions and quantify the uncertainty given noisy data. Prior probability distributions are placed over weights and the inference is then performed on weights, while the inference is not tractable in general. Therefore, variational inference is often used to approximate the inference and these models require more parameters and more computational cost to converge. To this end, dropout is utilized to estimate the uncertainty in \cite{zhang2019quantifying}, where arbitrary polynomial chaos is combined with PINN to solve stochastic PDE problems. In current researches, GAN, Bayesian method or dropout is integrated with PINN to quantify prediction uncertainty, where PINN is the basic network structure. PINN is particularly promising for ill-posed forward and inverse problems, while traditional numerical methods, such as finite element methods \cite{reddy2019introduction} and finite difference methods \cite{leveque2007finite}, require well-defined initial and boundary conditions, which is unavailable in most applications. However, PINN is much slower than traditional numerical methods \cite{bandai2022forward}. Therefore, how to combine the advantages of PINN and achieve numerical equivalent computational costs to quantification uncertainty is a critical challenge for solving PDE-based problems.

Recently, a novel NN called extreme learning machine (ELM) has received great attention owing to its low computational cost and simplicity \cite{yang2018novel,sun2019solving}. The most peculiar property of the ELM is that its input layer weights are pre-set random values and fixed throughout the training process, and output layer weights are training parameters. Spirited by ELM and PINN, Dwivedi et al. \cite{dwivedi2020physics,dwivedi2021distributed} develop an ELM-based PDE solver called physic-informed extreme learning machine (PIELM) as a rapid version of PINN to solve the linear PDE problems efficiently.
The PDE problem is transformed into a linear least squares problem by incorporating physics laws into the ELM as the cost function. Subsequently, the input layer weights are set randomly from a range according to the number of neurons, which is effective to solve elliptic PDEs with sharp gradients \cite{calabro2021extreme}. Besides, for nonlinear PDE problems, ELM-based PDE solvers transform the PDE problems into nonlinear least squares problems \cite{dong2021local,schiassi2021extreme}, where the NN is also trained by a nonlinear least squares computation, not by gradients-based algorithms. The ELM-based PDE solver as a data-driven technology inevitably faces noise scenarios in real-life applications. The ELM-based PDE solver, however, has two evident drawbacks for ill-posed problems with noisy data:
\begin{enumerate}[leftmargin=*]    
\item For linear PDE problems, the output weights solved by Moore–Penrose generalized inverse easily suffer from overfitting noisy data and the accuracy of the solver is drastically sensitive to the number of hidden neurons under noise scenarios.
\item The ELM-based PDE solvers are not equipped with the built-in uncertainty quantification at current works, which may restrict their real applications.       
\end{enumerate}

In this paper, we propose a novel Bayesian physics-informed extreme learning machine (BPIELM) to solve both forward and inverse PDE problems with noisy data, see Fig.\ref{fig:PINN_TFI}, where the Bayesian method is used to quantify the uncertainty from the scattered noisy data. The Bayesian framework naturally quantifies the uncertainty arising from noisy data \cite{luo2020bayesian}. BPIELM consists of two parts: the prior for the extreme learning machine with physics laws and the Bayesian method for estimating posterior distributions of parameters. The first part is to incorporate physics laws into the ELM as the cost function and introduce a prior distribution in the output layer weights, then the PDE problem is transformed into a linear least squares problem. In particular, for some inverse problems, the PDE problem parameters are considered as new output layer weights. The second part is to estimate posterior distributions of parameters and quantify the prediction uncertainty for the solution. We demonstrate the BPIELM method over a range of forward as well as inverse PDE problems and conduct a systematic comparison with PIELM and PINN. In summary, the followings are main benefits of BPIELM,
\begin{enumerate}[leftmargin=*]
\item BPIELM is equipped with the built-in uncertainty quantification, which provides more accurate predictions than PIELM under noise scenarios and has lower computational costs than PINN.
\item Compared to PIELM, BPIELM avoids overfitting by estimating the probability distribution of output values instead of fitting noisy data, and hence is not drastically sensitive to the number of hidden neurons.
\item BPIELM is also used for some inverse problems in a unified framework, which achieves a competitive prediction accuracy with PIELM and PINN under noise scenarios.
\end{enumerate}

The rest of the paper is structured as follows. Problem statement is presented in section \ref{sec problem}. In section \ref{sec method}, we first briefly review PINN and PIELM, and then present the BPIELM method in detail. Numerical analysis is conducted in section \ref{sec result}, where we discuss the results of BPIELM, PIELM and PINN for forward and inverse PDE problems. The conclusions of the paper are in section \ref{sec conclusion}.

\section{Problem Setup}
\label{sec problem}
Suppose we have a linear partial differential equation (PDE) describing a physical system as follows:

\begin{equation}
\label{eq_problem_set}
\begin{array}{l}
Lu=f_{\lambda}(x,y),(x,y)\in \Omega, \\
\mathcal{B}[u(x,y)]=b(x,y),(x,y)\in\partial \Omega,
\end{array}
\end{equation}
where $L$ is a linear differential operator, $\mathcal{B}$ is a boundary/initial condition operator acting on the boundary $\partial \Omega$, while $u(x,y)$ and $f_{\lambda}(x,y)$ are prescribed the solution and source terms parameterized by the problem parameter $\lambda$, respectively. 

In this work, we consider two types of data-driven problems, as summarized in Table \ref{tab_problem} and delineated below (refer to \cite{psaros2022uncertainty} for more problems). The first scenario pertains to a forward PDE problem, where $L$ and $\mathcal{B}$ are known. The problem parameter $\lambda$ is also known and the solution $u(x,y)$ is unknown. Given noisy data of $b$ sampled on $\partial \Omega$, our quantity of interest is to infer the solution $u(x,y)$ at every $(x,y)\in \Omega$ and quantify its uncertainty. Assume we only have $N_{b}$ sensors on the boundary (including sensors of initial condition for simplification) and the dataset in this scenario is described as $\mathcal{D}=\left\{\mathcal{D}_{b}\right\}$, $\mathcal{D}_{b}=\left\{(x_{b}^{(i)},y_{b}^{(i)}), \bar{b}^{(i)}\right\}_{i=1}^{N_{b}}$. Assume that the measurements are independently Gaussian distributed centered at the hidden real value as follows:
\begin{equation}
\label{eq_f_noisy}
\begin{array}{ll}
\bar{b}^{(i)}=b\left(x_{b}^{(i)},y_{b}^{(i)}\right)+\epsilon_{b}^{(i)}, & i=1,2 \ldots N_{b},
\end{array}
\end{equation}
where $\epsilon_{b}^{(i)}$ is independent Gaussian noises with zero mean. Assume that the fidelity of each sensor is known and the standard deviation of $\epsilon_{b}^{(i)}$ is known to be $\sigma_{b}^{(i)}$.

The second scenario pertains to an inverse PDE problem, where $L$ and $\mathcal{B}$ are known. The problem parameter $\lambda$ is unknown, while $u(x,y)$ is partially known and some extra measurements on $u(x,y)$ are available. Given noisy data of $u$ in $\Omega$ as well as noisy data of $b$, our quantity of interest is to infer not only the solution $u(x,y)$ at every $(x,y)\in \Omega$ but the problem parameter $\lambda$, and quantify its uncertainty. Assume we have $N_{u}$ sensors in the domain $\Omega$ and $N_{b}$ sensors on the boundary $\partial \Omega$. The dataset in this scenario is described as $\mathcal{D}=\left\{\mathcal{D}_{u},\mathcal{D}_{b} \right\}$, where $\mathcal{D}_{u}=\left\{(x_{u}^{(i)},y_{u}^{(i)}), \bar{u}^{(i)}\right\}_{i=1}^{N_{u}}$ and $\mathcal{D}_{b}=\left\{(x_{b}^{(i)},y_{b}^{(i)}), \bar{b}^{(i)}\right\}_{i=1}^{N_{b}}$. Similarly, the measurements from $\mathcal{D}_{u}$ and $\mathcal{D}_{b}$ are represented as


\begin{equation}
\label{eq_u_noisy}
\begin{array}{ll}
\bar{u}^{(i)}=u\left(x_{u}^{(i)},y_{u}^{(i)}\right)+\epsilon_{u}^{(i)}, i=1,2 \ldots N_{u},\\
\bar{b}^{(i)}=b\left(x_{b}^{(i)},y_{b}^{(i)}\right)+\epsilon_{b}^{(i)}, i=1,2 \ldots N_{b},
\end{array}
\end{equation}
where $\epsilon_{u}^{(i)}$ is also the independent Gaussian noise with zero mean. Assume that the fidelity of each sensor is known and the standard deviation of $\epsilon_{u}^{(i)}$ is known to be $\sigma_{u}^{(i)}$.

\begin{table*}[!http]
\centering
\caption{Problem  considered in this paper.}
\label{tab_problem}
\scalebox{0.8}{
\begin{tabular}{rlllllll}
\hline
\multicolumn{8}{c}{\textbf{Considered data-driven problem corresponding to following equation}}
\\
\multicolumn{8}{c}{$Lu=f_{\lambda}(x,y),(x,y)\in \Omega,$ and $ \mathcal{B}[u(x,y)]=b(x,y),(x,y)\in\partial \Omega$}
\\ \hline
\multicolumn{8}{c}{\textit{\textbf{Forward PDE problem}}}                                           \\ \hline
\multicolumn{1}{r|}{$L$,$\mathcal{B}$}          & \multicolumn{3}{l|}{known}                                                                               & \multicolumn{4}{c}{\multirow{7}{*}{\textbf{}}}  \\
\multicolumn{1}{r|}{$f_{\lambda}$,$b$}          & \multicolumn{3}{l|}{noisy data}                                                                         & \multicolumn{4}{c}{} \multirowcell{4}{\includegraphics[scale=0.4]{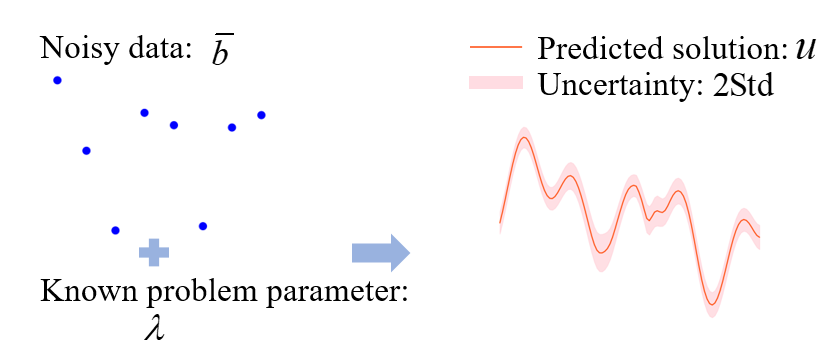}}                          \\
\multicolumn{1}{r|}{$u(x,y)$}                   & \multicolumn{3}{l|}{unknown}                                                                             & \multicolumn{4}{c}{}                           \\
\multicolumn{1}{r|}{$\lambda$}                  & \multicolumn{3}{l|}{known}                                                                                           & \multicolumn{4}{c}{}                           \\
\multicolumn{1}{r|}{Dataset}                    & \multicolumn{3}{l|}{$\mathcal{D}=\left\{ \mathcal{D}_{b}\right\}$,where}                                      & \multicolumn{4}{c}{}                           \\
\multicolumn{1}{l|}{}                           & \multicolumn{3}{l|}{$\mathcal{D}_{b}=\left\{(x_{b}^{(i)},y_{b}^{(i)}), \bar{b}^{(i)}\right\}_{i=1}^{N_{b}}$} & \multicolumn{4}{c}{}                           \\
\multicolumn{1}{c|}{Objective}                  & \multicolumn{3}{l|}{$u(x,y)$, uncertainty}                               & \multicolumn{4}{c}{}                           \\ \hline
\multicolumn{8}{c}{}                                                     \\ \hline
\multicolumn{8}{c}{\textit{\textbf{Inverse PDE problem}}}                                                                           \\ \hline
\multicolumn{1}{r|}{$L$,$\mathcal{B}$}          & \multicolumn{3}{l|}{known}                                                                                  & \multicolumn{4}{l}{\multirow{8}{*}{}}        \\
\multicolumn{1}{r|}{$f_{\lambda}$, $b$}         & \multicolumn{3}{l|}{noisy data}                                                                                                   & \multicolumn{4}{l}{}  \multirowcell{4}{\includegraphics[scale=0.39]{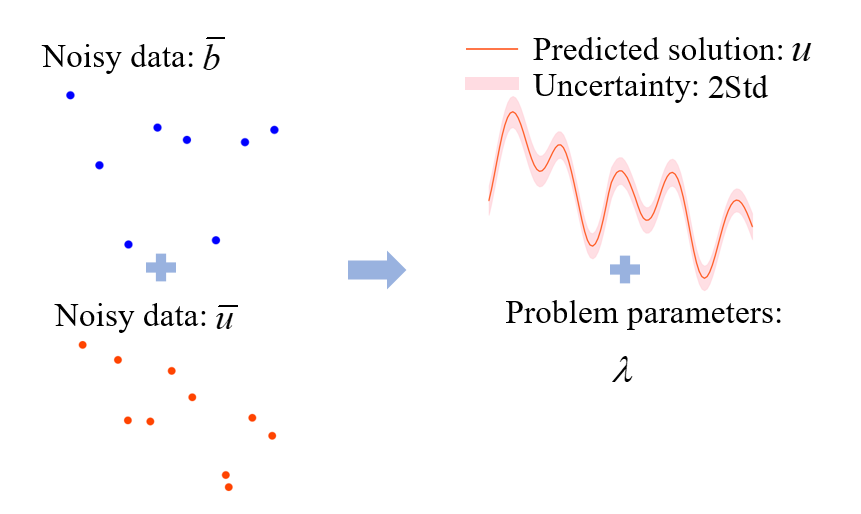}}                           \\
\multicolumn{1}{r|}{$u(x,y)$}                   & \multicolumn{3}{l|}{partially known}                                                                                            & \multicolumn{4}{l}{}                           \\
\multicolumn{1}{r|}{$\lambda$}                  & \multicolumn{3}{l|}{unknown}                                                                                             & \multicolumn{4}{l}{}                           \\
\multicolumn{1}{r|}{Dataset}                    & \multicolumn{3}{l|}{$\mathcal{D}=\left\{\mathcal{D}_{u}, \mathcal{D}_{b}\right\}$,where}                                                                                                       & \multicolumn{4}{l}{}                           \\
\multicolumn{1}{l|}{}                           & \multicolumn{3}{l|}{$\mathcal{D}_{u}=\left\{(x_{u}^{(i)},y_{u}^{(i)}), \bar{u}_{i}\right\}_{i=1}^{N_{u}}$,$\mathcal{D}_{b}=\left\{(x_{b}^{(i)},y_{b}^{(i)}), \bar{b}_{i}\right\}_{i=1}^{N_{b}}$}    & \multicolumn{4}{l}{}                           \\
\multicolumn{1}{l|}{}                           & \multicolumn{3}{l|}{}                                                                                                       & \multicolumn{4}{l}{}                           \\
\multicolumn{1}{c|}{Objective}                  & \multicolumn{3}{l|}{$\lambda$, $u(x,y)$, uncertainty}                                                                                                                                 & \multicolumn{4}{l}{}                           \\ \hline
\end{tabular}
}
\end{table*}

\section{Method}\label{sec method}

\subsection{A primer in physics-informed neural network and physics-informed extreme learning machine}
PIELM is considered as a rapid version of PINN, while BPIELM is the combination of PIELM and the Bayesian methods. Therefore, in this subsection, we briefly review PINN and PIELM.
\subsubsection{Physics-informed neural network}

PINN proposed by Raissi et al. \cite{raissi2019physics} is utilized to solve forward and inverse PDE problems. The key idea of PINN is to encode the physic laws into loss function by adding a penalty term to constrain the space of admissible solutions and then the problem of solving PDEs in Eq.(\ref{eq_problem_set}) is transformed into an optimization problem of minimizing the
loss function. Concretely, a multi-layer perception with the activation function is first constructed. Then, to measure the difference between the NN and physics laws, the loss function is defined as
\begin{equation}
\mathcal{L}(\theta)=w_{f} \mathcal{L}_{f}\left(\theta ; \mathcal{T}_{f}\right)+w_{b} \mathcal{L}_{b}\left(\theta ; \mathcal{T}_{b}\right)+w_{u} \mathcal{L}_{\text {u }}\left(\theta ; \mathcal{T}_{\text {u }}\right),
\end{equation}
where 
\begin{equation}
\begin{aligned}
\mathcal{L}_{f}\left(\theta ; \mathcal{T}_{f}\right) &=\frac{1}{\left|\mathcal{T}_{f}\right|} \sum_{\mathbf{x} \in \mathcal{T}_{f}}\|f(t, \mathbf{x})\|^{2}, \\
\mathcal{L}_{b}\left(\theta ; \mathcal{T}_{b}\right) &=\frac{1}{\left|\mathcal{T}_{b}\right|} \sum_{\mathbf{x} \in \mathcal{T}_{b}}\|\mathcal{B}(\mathbf{x})-b\|^{2}, \\
\mathcal{L}_{\text {u }}\left(\theta ; \mathcal{T}_{\text {u }}\right) &=\frac{1}{\left|\mathcal{T}_{\text {u }}\right|} \sum_{\mathbf{x} \in \mathcal{T}_{\text {u }}}\|\hat{u}(\mathbf{x})-u(\mathbf{x})\|^{2}.
\end{aligned}
\end{equation}
$w_{f}, w_{b}$ as well as $w_{u}$ are hyper-parameters of weights. $f$ represents the PDE residual in Eq.(\ref{eq_problem_set}). $\theta$ represents weights and bias of the NN. $\mathcal{T}_{f}$, $\mathcal{T}_{b}$, and $\mathcal{T}_{u}$ denote training points from PDE residual, boundary and data constraints, respectively. In other word, the loss function of PINN is a weighted sum of multiple terms, including PDE, boundary and data constraints. 

Finally the NN is trained to search for the best parameters $\theta$ by minimizing the loss function, where automatic differentiation is used to minimize the loss function.
\subsubsection{Physics-informed extreme learning machine}
Spirited by PINN, Dwivedi et al. \cite{dwivedi2020physics} propose a rapid version of PINN called PIELM, which can be applied to stationary and time-dependent linear PDEs. The most peculiar property is that PIELM is a single-layer feed-forward NN, where
its input layer weights are fixed with random values and only output layer weights are training parameters. Then, the physics laws are incorporated into the NN as the cost function. In particular, the linear nature of linear PDEs and the physics laws in Eq.(\ref{eq_problem_set}) are combined to solve a linear equation as follows:
\begin{equation}
\label{eq_PIELM}
\begin{array}{l}
\mathbf{H}\boldsymbol{c}=\mathbf{T},
\end{array}
\end{equation}
where $\boldsymbol{c}$ is the unknown output layer weights and $\mathbf{H}$ as well as $\mathbf{T}$ are known matrices that contain physic laws (PDE, boundary and data constraints). Finally, the predictions of $\boldsymbol{c}$ come from linear least squares solution by
\begin{equation}
\label{eq_PIELM_inver}
\begin{array}{l}
\boldsymbol{c}_{pred}=\mathbf{H}^{\dagger}\mathbf{T},
\end{array}
\end{equation}
where $\mathbf{H}^{\dagger}=\left(\mathbf{H}^{T} \mathbf{H}\right)^{-1} \mathbf{H}^{T}$ represents the Moore–Penrose generalized inverse matrix of $\mathbf{H}$. 

\subsection{Bayesian physics-informed extreme learning machine}
\label{sec BPIELM}
\begin{figure*}[htbp]
	\centering
	\includegraphics[width=0.8\linewidth]{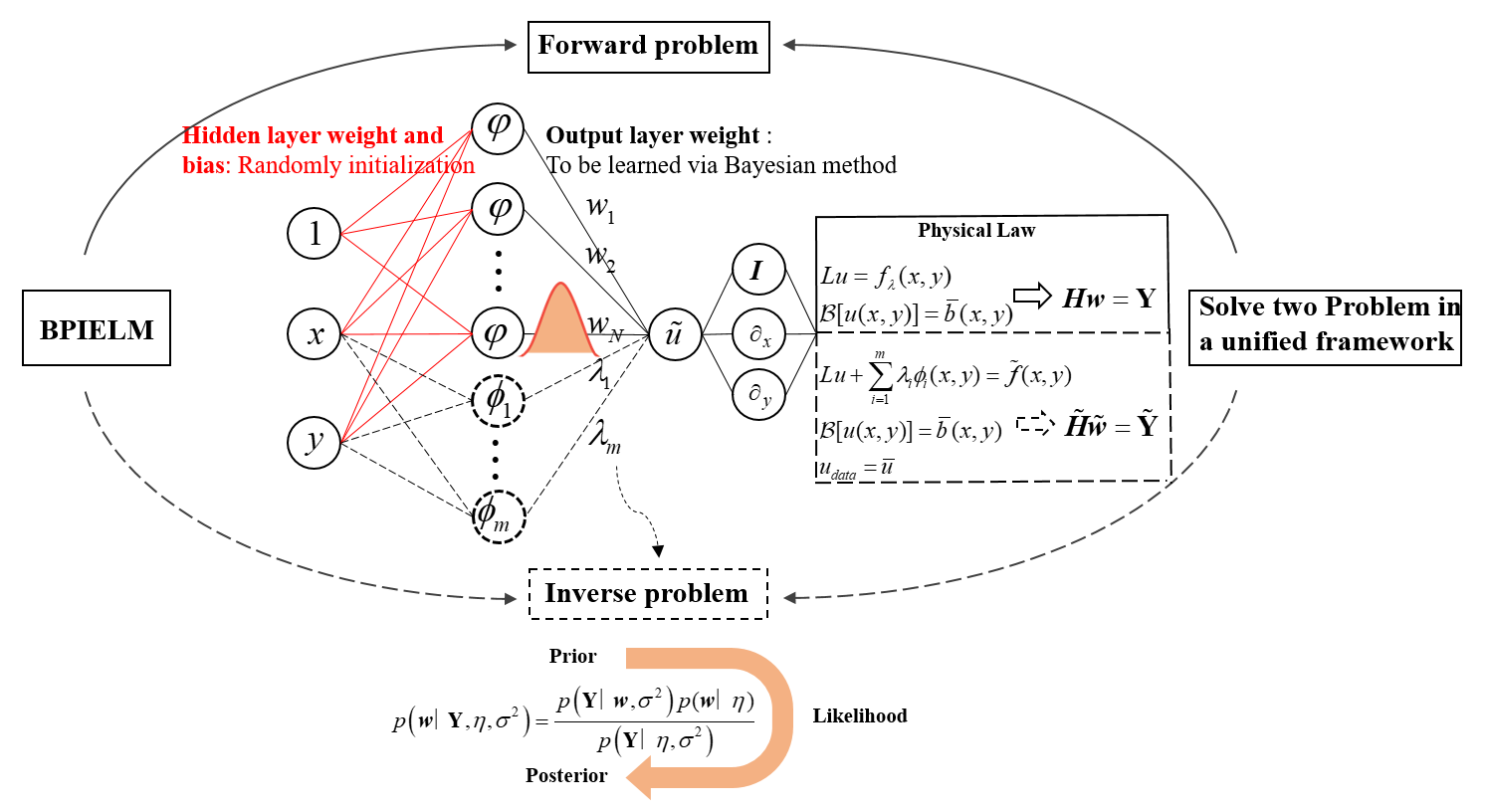}
	\caption{Conceptual flow of the BPIELM method with noisy data. $p(\boldsymbol{\omega} \mid \eta)$ is the prior for the output layer weights as well as problem parameters, $p\left(\mathbf{Y} \mid \boldsymbol{\omega}, \sigma^{2}\right)$ represents the likelihood of measurements, and $p\left(\boldsymbol{\omega} \mid \mathbf{Y},\eta, \sigma^{2}\right)$ represents the posterior.}
	\label{fig:PINN_TFI}
\end{figure*}

BPIELM is an ELM-based PDE solver that combines two ideas from PIELM and the Bayesian method. The first idea, coming from PIELM, is to incorporate the physics laws into ELM as the cost function, making linear PDE problems transformed into a linear least squares problem. The other one, coming from Bayesian ELM \cite{soria2011belm, jin2020eeg}, is combining the Bayesian method with ELM to naturally quantify the uncertainty for regression and classification tasks. 
The output layer parameters are defined as a prior distribution and are optimized by the Bayesian method, not by the Moore–Penrose generalized inverse. Since PIELM is not equipped with the built-in uncertainty quantification and is prone to overfit noisy data, we resolve two drawbacks by combining characteristics between PIELM and the Bayesian framework. In addition, we also unify the forward and inverse PDE problems in the one framework. BPIELM consists of two parts: the prior for extreme learning machine with physics laws and the Bayesian method for estimating posterior distributions of parameters. The first part is incorporating physics laws into the NN as the cost function and introducing a prior in the output layer weights, then the PDE problem is transformed into a linear least squares problem. The second part is to estimate posterior distributions of parameters and quantify the prediction uncertainty. Fig.\ref{fig:PINN_TFI} shows the conceptual flow of the BPIELM method and the detailed descriptions are provided in the following sections.

\subsubsection{The prior for extreme learning machine with physics laws}
We consider a single-layer NN with $N$ neurons in Fig.\ref{fig:PINN_TFI}. The key of the ELM is to pre-set the input layer weights randomly, which are fixed throughout the training process. The parameters of the input layer are defined as $\boldsymbol{\chi}=[x, y, 1]^{T}$, $\boldsymbol{\alpha}=[\alpha_{1}, \alpha_{2}, \cdots, \alpha_{n}]^{T}$, $\boldsymbol{\beta}=[\beta_{1}, \beta_{2}, \cdots, \beta_{n}]^{T}$,  $\boldsymbol{\gamma}=[\gamma_{1}, \gamma_{2}, \cdots, \gamma_{n}]^{T}$, where $\boldsymbol{\alpha}$, $\boldsymbol{\beta}$ and $\boldsymbol{\gamma}$ are input layer weights. The nonlinear activation function is $\varphi=\tanh$ and the output layer weights are $\boldsymbol{\omega}=[\omega_{1}, \omega_{2}, \cdots, \omega_{n}]^{T}$. Then the output of $k^{th}$ hidden neuron is represented by $\varphi\left(z_{k}\right)$, where $z_{k}=[\alpha_{k},\beta_{k},\gamma_{k}]\boldsymbol{\chi}$. The output of the NN is written as 
\begin{equation}
\label{eq_u_NN}
\hat{u}(\boldsymbol{\chi})=\varphi\left(\boldsymbol{z}\right) \boldsymbol{\omega}.
\end{equation}
Given $\hat{u}(\boldsymbol{\chi})$, the partial derivatives of Eq.(\ref{eq_u_NN}), with respect to the independent variables, are
\begin{equation}
\begin{array}{l}
\frac{\partial^{k} \hat{u}}{\partial x^{k}}=\sum_{j=1}^{n} \omega_{j} \alpha_{j}^{k} \frac{\partial^{k} \varphi}{\partial z^{k}}, \\
\frac{\partial^{k} \hat{u}}{\partial y^{k}}=\sum_{j=1}^{n} \omega_{j} \beta_{j}^{k} \frac{\partial^{k} \varphi}{\partial z^{k}},
\end{array}
\end{equation}
where $k$ refers to the $k^{th}$ order derivative.

Therefore, according to Eq.(\ref{eq_problem_set}), (\ref{eq_f_noisy}) and (\ref{eq_u_noisy}), the physics laws are incorporated into the NN by introducing physics-based training errors, $\boldsymbol{\xi}_{f}, \boldsymbol{\xi}_{b}$. The training errors $\boldsymbol{\xi}_{f}, \boldsymbol{\xi}_{b}$ are defined as errors in approximating PDE and boundary/initial condition, respectively,
\begin{equation}
\label{eq_pde_f}
\begin{array}{l}
\boldsymbol{\xi}_{f} = L\varphi\left(\boldsymbol{z}\right) \boldsymbol{\omega}-f_{\lambda}(x,y)=0, \\
\boldsymbol{\xi}_{b} = \mathcal{B}[\varphi\left(\boldsymbol{z}\right) \boldsymbol{\omega}]-\bar{b}(x,y)=0.
\end{array}
\end{equation}

For the forward problems, Eq.(\ref{eq_pde_f}) leads to a linear equation, which is represented as 
\begin{equation}
\label{eq_linear_forward}
\begin{array}{l}
\mathbf{H}\boldsymbol{\omega}=\mathbf{Y},
\end{array}
\end{equation}
where $\mathbf{H} \in \mathbb{R}^{N^{*} \times N}$, $N^{*}=N_{f}+N_{b}$, N is equal to the neuron number. $\mathbf{H}$ and $\mathbf{Y}$ are matrices determined by $L$ and $\mathcal{B}$. The output layer weight $\boldsymbol{w}$ is the training parameters.

In particular, for some inverse PDE problems in this work, we consider the type of PDEs and its training errors in approximating the PDE can be written by separating variables as
\begin{equation}
\label{eq_pde_inverse_f}
\begin{array}{l}
\boldsymbol{\xi}_{f} = L\varphi\left(\boldsymbol{z}\right) \boldsymbol{\omega}+\sum_{j=1}^{m}\phi_{j}\lambda_{j}-\tilde{f}=0,
\end{array}
\end{equation}
where $\phi_{j}$ and $\tilde{f}$ are determined by separating variables. Under noisy data in Eq.(\ref{eq_u_noisy}), the training errors in approximating the boundary/initial condition and data constraints are represented by

\begin{equation}
\label{eq_u_inverse}
\begin{array}{l}
\boldsymbol{\xi}_{b}=\left(H_{b}, 0\right)\left(\begin{array}{c}
\boldsymbol{\omega}, \\
\boldsymbol{\lambda},
\end{array}\right)-\bar{b}(x, y)=0, \\
\boldsymbol{\xi}_{u}=\left(H_{u}, 0\right)\left(\begin{array}{c}
\boldsymbol{\omega} \\
\boldsymbol{\lambda}
\end{array}\right)-\bar{u}(x, y)=0.
\end{array}
\end{equation}
For the inverse problems, Eq.(\ref{eq_pde_inverse_f}) and (\ref{eq_u_inverse}) also lead to a linear system,
\begin{equation}
\label{eq_linear_inverse}
\begin{array}{l}
\mathbf{\tilde{H}}\boldsymbol{\tilde{\omega}}=\mathbf{\tilde{Y}},
\end{array}
\end{equation}
where $\boldsymbol{\tilde{\omega}}=[\boldsymbol{\omega}, \boldsymbol{\lambda}]^{T}$, $\mathbf{\tilde{H}}\in \mathbb{R}^{N^{*} \times N},N^{*}=N_{f}+N_{b}+N_{u}$ as well as $\mathbf{\tilde{Y}}$, are matrices determined by the $L$ and $\mathcal{B}$. The problem parameters are also considered as output layer weights. 

In summary, physics laws are encoded into the linear system by using the linear nature of linear PDEs. To simplify the following representation, the linear system for forward or inverse problems is written as $\mathbf{H}\boldsymbol{\omega}=\mathbf{Y}$, where $\boldsymbol{\omega}$ is the training parameters. In most applications of Bayesian theory, a commonly used prior for $\boldsymbol{\omega}$ is that each component of $\boldsymbol{\omega}$ is an independent Gaussian distribution with zero mean, which is represented by
\begin{equation}
p(\boldsymbol{\omega} \mid \eta)=\mathcal{N}\left(\mathbf{0} ; \eta^{-1} \mathbf{I}\right),
\end{equation}
where $\mathbf{I}$ is the identity matrix and $\eta$ is a hyperparameter, which will be optimized the following steps.


\subsubsection{The Bayesian method for estimating posterior distributions of parameters}
In this subsection, the Bayesian method is used to estimate posterior distributions of parameters, which improves the over-fitting problem of ELM-based solvers and quantifies the prediction uncertainty. According to wide application of Bayesian theory, most models are divided into two steps:

\begin{enumerate}[leftmargin=*]     
\item Inference of the posterior distribution of model parameters. Suppose $\boldsymbol{\omega}$ is the set of parameters and $\mathcal{D}$ is the dataset. The posterior distribution is proportional to the product of the prior
distribution and the likelihood function,
\begin{equation}
\begin{array}{l}
P(\boldsymbol{\omega} \mid D) \propto P(\boldsymbol{\omega}) P(D \mid \boldsymbol{\omega}).
\end{array}
\end{equation}
\item The output $\mathbf{Y}_{new}$ can be given by the
integral of the posterior distribution of $\boldsymbol{\omega}$ for input instances of $\boldsymbol{\chi}_{new}$,

\begin{equation}
\label{eq_posterior_distribution}
P\left(\mathbf{Y}_{\boldsymbol{new}} \mid \boldsymbol{\chi}_{\boldsymbol{new}}, \boldsymbol{D}\right)=\int P\left(\mathbf{Y}_{\boldsymbol{new}} \mid \boldsymbol{\chi}_{\boldsymbol{new}}, \boldsymbol{\omega}\right) P(\boldsymbol{\omega} \mid \boldsymbol{D}) d \boldsymbol{\omega}.
\end{equation}
\end{enumerate}

Suppose the linear system follows the relationship,
\begin{equation}
\mathbf{Y}=\mathbf{H} \boldsymbol{\omega}+\epsilon,
\end{equation}
where $\epsilon$ follows a Gaussian distribution $\mathcal{N}\left(0 ; \sigma^{2}\right)$. The probability model and likelihood function are then derived as
\begin{equation}
P\left(\mathbf{Y} \mid \boldsymbol{\omega}, \sigma^{2}\right) \\
=\left(\frac{1}{2 \pi \sigma^{2}}\right)^{\frac{N^{*}}{2}} \exp \left(-\frac{\|\mathbf{Y}-\mathbf{H} \boldsymbol{\omega}\|^{2}}{\sigma^{2}}\right),
\end{equation}
where $N^{*}$ is the dimension number of $\mathbf{H}$.

Under the assumption that the prior distribution and the likelihood function follow Gaussian distributions, the posterior can be derived as
\begin{equation}
p\left(\boldsymbol{\omega} \mid \mathbf{Y},\eta, \sigma^{2}\right)=\frac{p\left(\mathbf{Y} \mid \boldsymbol{\omega}, \sigma^{2}\right) p(\boldsymbol{\omega} \mid \eta)}{p\left(\mathbf{Y} \mid \eta, \sigma^{2}\right)},
\end{equation}
which also follows a Gaussian distribution with a mean value of $\mu$ and a variance of $\boldsymbol{\Sigma}$ \cite{chen2009bayesian}, 

\begin{equation}
\begin{array}{c}
\boldsymbol{\mu}=\sigma^{-2} \boldsymbol{\Sigma} \boldsymbol{H}^{\boldsymbol{T}} \mathbf{Y},\\
\boldsymbol{\Sigma}=\left(\eta \boldsymbol{I}+\sigma^{-2} \boldsymbol{H}^{\boldsymbol{T}} \boldsymbol{H}\right)^{-1}.
\end{array}
\end{equation}

Then the optimal values of hyper-parameters, $\eta$ and $\sigma$, are obtained by Evidence Procedure \cite{mackay1995probable}. The optimal conditions are iteratively calculated by
\begin{equation}
\begin{array}{c}
\eta \leftarrow \frac{N-\eta \cdot trace[\boldsymbol{\mu}]}{\boldsymbol{\Sigma}^{\boldsymbol{T}} \boldsymbol{\Sigma}},\\
\sigma^{2} \leftarrow \frac{\|\mathbf{Y}-\mathbf{H} \boldsymbol{\omega}\|^{2}}{N^{*}-N+\eta \cdot trace[\boldsymbol{\mu}]},
\end{array}
\end{equation}
where $N^{*}$ is the dimension number of $\mathbf{H}$ and $N$ is the number of parameters. The iterative process stops when $\boldsymbol{\mu}$ falls below a given threshold.

Finally, given a $\boldsymbol{\chi}_{\boldsymbol{new}}$, the posterior distributions of the parameters are used by Eq.(\ref{eq_posterior_distribution}) to obtain the output $\mathbf{Y}_{\boldsymbol{new}}$. Besides, the output follows a distribution,

\begin{equation}
p\left(\mathbf{Y}_{\boldsymbol{new}} \mid \mathbf{Y}, \eta, \sigma^{2}\right)=\mathcal{N}\left(\boldsymbol{H}_{\boldsymbol{new}} \boldsymbol{\omega}; \sigma^{2}\left(\boldsymbol{\chi}_{\boldsymbol{new}}\right)\right),
\end{equation}
where $\boldsymbol{H}_{\boldsymbol{new}}=\varphi ([\boldsymbol{\alpha}, \boldsymbol{\beta}, \boldsymbol{\gamma}]\chi_{\boldsymbol{new}} )$ and the variance is calculated by

\begin{equation}
\sigma^{2}\left(\boldsymbol{\chi}_{new }\right)=\sigma^{2}+\boldsymbol{\chi}_{\boldsymbol{new}}^{T} \boldsymbol{\Sigma} \boldsymbol{\chi}_{\boldsymbol{new}}^{T}.
\end{equation}

\section{Numerical result}
\label{sec result}
In this section, we will first consider solving forward and inverse PDE problems by BPIELM over PIELM and PINN. The forward PDE problem including Poisson, advection and diffusion equations. The inverse PDE problem includes Poisson and Helmholtz equations, where noisy data is used to solve the solution and identify the unknown problem parameters. 
All the experiments are conducted the very common machine with a 2.8-GHz Intel(R) Xeon(R) Gold 6242 CPU, 128-GB memory, and NVIDIA GeForce RTX 3090 GPU.

\subsection{Poisson equation}
The two dimensional (2D) Poisson equation is given by 
\begin{equation}
\label{eq_poisson_2d}
u_{x x}+u_{y y}=\left(16 x^{2}+64 y^{2}-12\right) e^{-\left(2 x^{2}+4 y^{2}\right)}, (x, y) \in \Omega,
\end{equation}
where the solution is $u=\frac{1}{2}+e^{-\left(2 x^{2}+4 y^{2}\right)}$, $\Omega=\{(x, y) \mid x=0.55 \rho(\theta) \cos (\theta), y=0.75 \rho(\theta) \sin (\theta)\}$, and $\rho(\theta)=1+\cos (\theta) \sin (4 \theta), 0 \leq \theta \leq 2 \pi$. Assuming that the boundary condition is unknown, we only have $N_{b}$ sensors on the boundary, which are equidistantly distributed on $\partial\Omega$, and consider the Gaussian noise in the measurements.

\begin{figure*}[!htbp]
	\centering
	\subfigure{
		\includegraphics[width=0.9\linewidth]{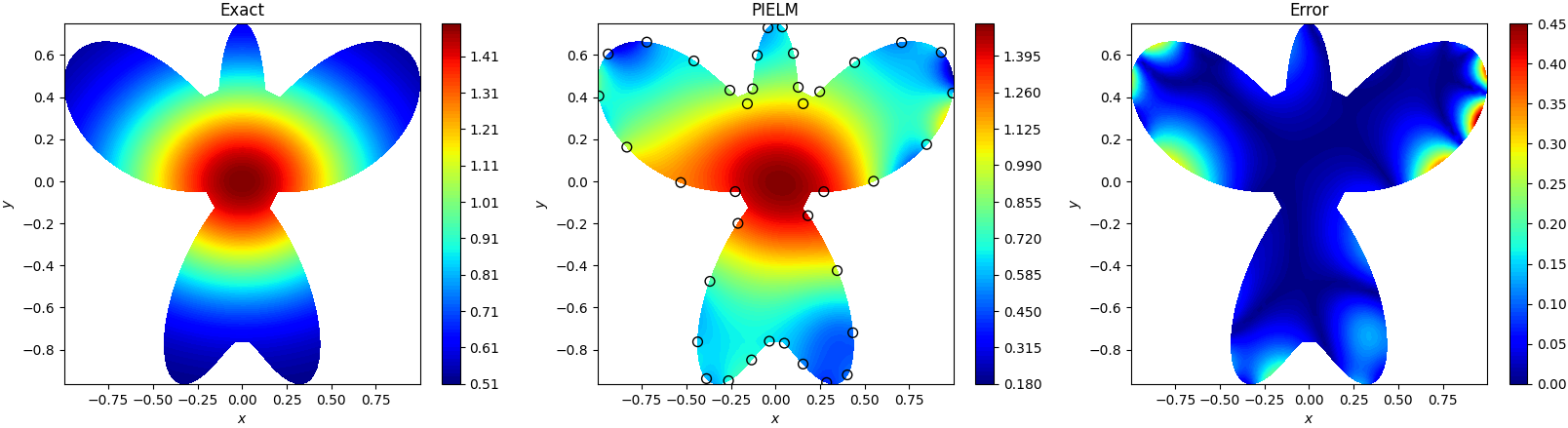}
	}
	\subfigure{
		\includegraphics[width=0.9\linewidth]{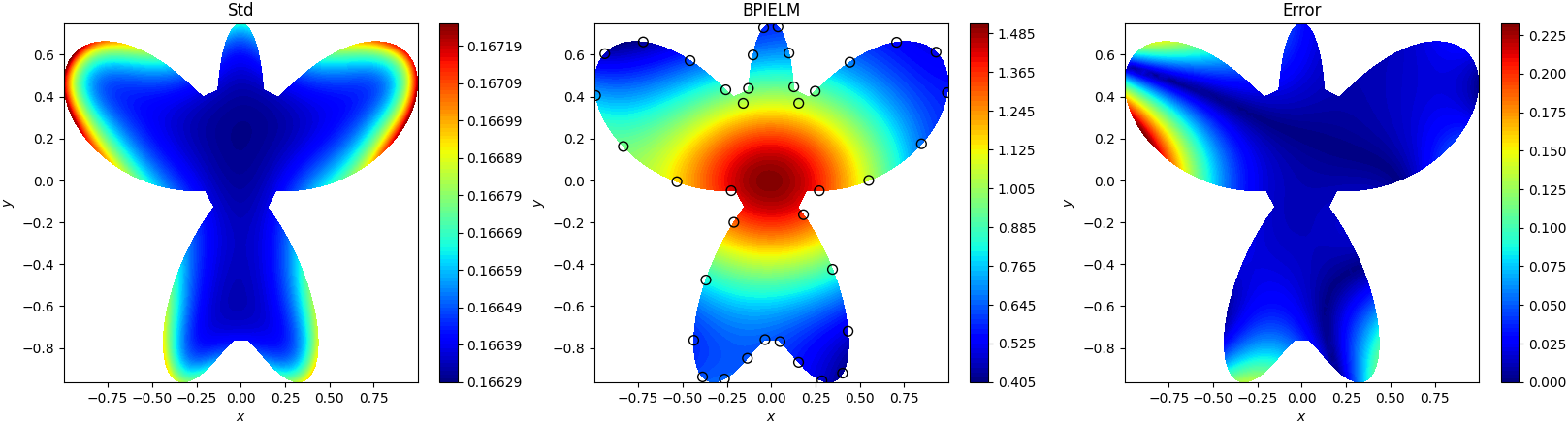}
	}
	\caption{Poisson equation: The first row represents the exact solution, the PIELM solution and its absolute error, respectively. The second row represents the standard deviations for u using BPIELM, the BPIELM solution and its absolute error, respectively. The noise scale on the boundary is $\epsilon_{b} \sim \mathcal{N}\left(0,0.01^{2}\right)$, the number of neurons is $N=100$, and the number of training points is $N_{f}=400$ and $N_{b}=38$. Black circles represent the positions of sensors.}
	\label{fig:Poisson_pred}
\end{figure*}

\begin{figure*}[htbp]
	\centering
	\includegraphics[width=1\linewidth]{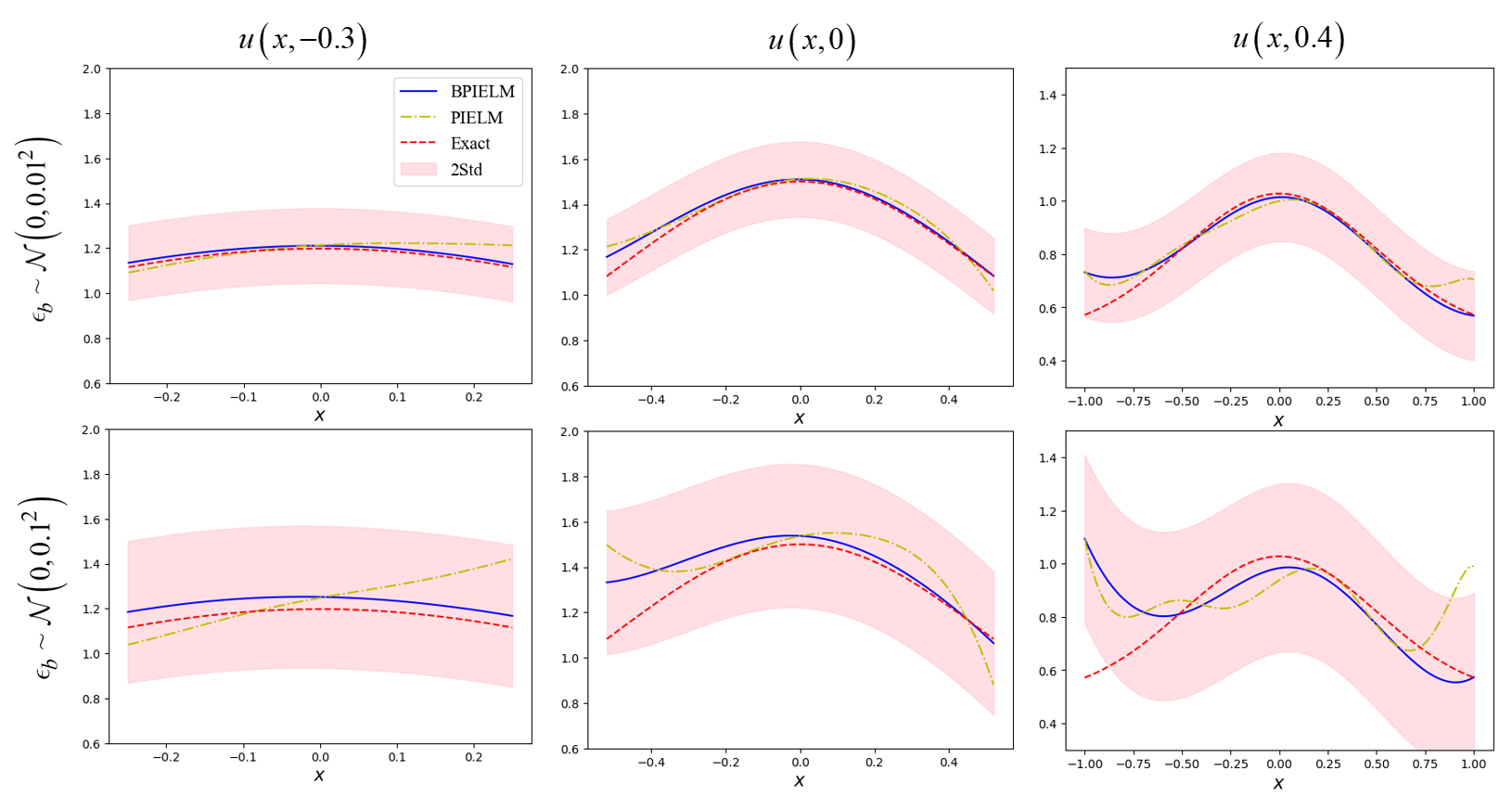}
	\caption{Poisson equation: The columns (from left to right) represent comparisons between BPIELM and PIELM at $y=-0.3$, $y=0$ and $y = 0.4$, where $N=100$, $N_{f}=400$ and $N_{b}=41$ are used. The row (from top to down) represents two noise scales on the boundary, $\epsilon_{b} \sim \mathcal{N}\left(0,0.01^{2}\right)$ and $\epsilon_{b} \sim \mathcal{N}\left(0,0.1^{2}\right)$.}
	\label{fig:Poisson_t}
\end{figure*}

To solve the Poisson equation in the BPIELM framework, we encode governing equations in Eq.(\ref{eq_poisson_2d}) and sensor measurements on the boundary into the linear system. Then, we use the Bayesian method to estimate the output layer weights, where the hyperparameters are $\eta=2e-1$ and $\beta=1$ (the default setting in the following tests). In this case, the number of neurons is $N=100$ and the number of training points is $N_{f}=400$ and $N_{b}=38$. The noise scale on the boundary is $\epsilon_{b} \sim \mathcal{N}\left(0,0.01^{2}\right)$. Fig.\ref{fig:Poisson_pred} shows the results of PIELM and BPIELM, where BPIELM provides more accurate predictions for the solution and gives the uncertainty quantification for the solution. In BPIELM, the standard deviation for the case is large on the boundary even though there are some sensors on the boundary. Besides, the area with a large error has a relatively large standard deviation. The mean absolute errors (MAEs) of PIELM and BPIELM are $0.029$ and $0.019$, respectively. The areas with large errors are mainly concentrated in the areas without sensors, while BPIELM gives predicted means closer to the exact solutions on the right-wing of the butterfly than PIELM. Fig.\ref{fig:Poisson_t} provides further comparisons between PIELM and BPIELM under two different noise scales, i.e., $\epsilon_{b} \sim \mathcal{N}\left(0,0.01^{2}\right)$ and $\epsilon_{b} \sim \mathcal{N}\left(0,0.1^{2}\right)$. BPIELM provides predictions closer to the exact solution than PIELM under two noise scales, and the errors of BPIELM are mostly bounded by two standard deviations, which increase as the noise scale increases. Moreover, the errors between predictions and the exact solution increase with the increasing noise scale.

\begin{figure*}[!t]
	\centering
	\includegraphics[width=1\linewidth]{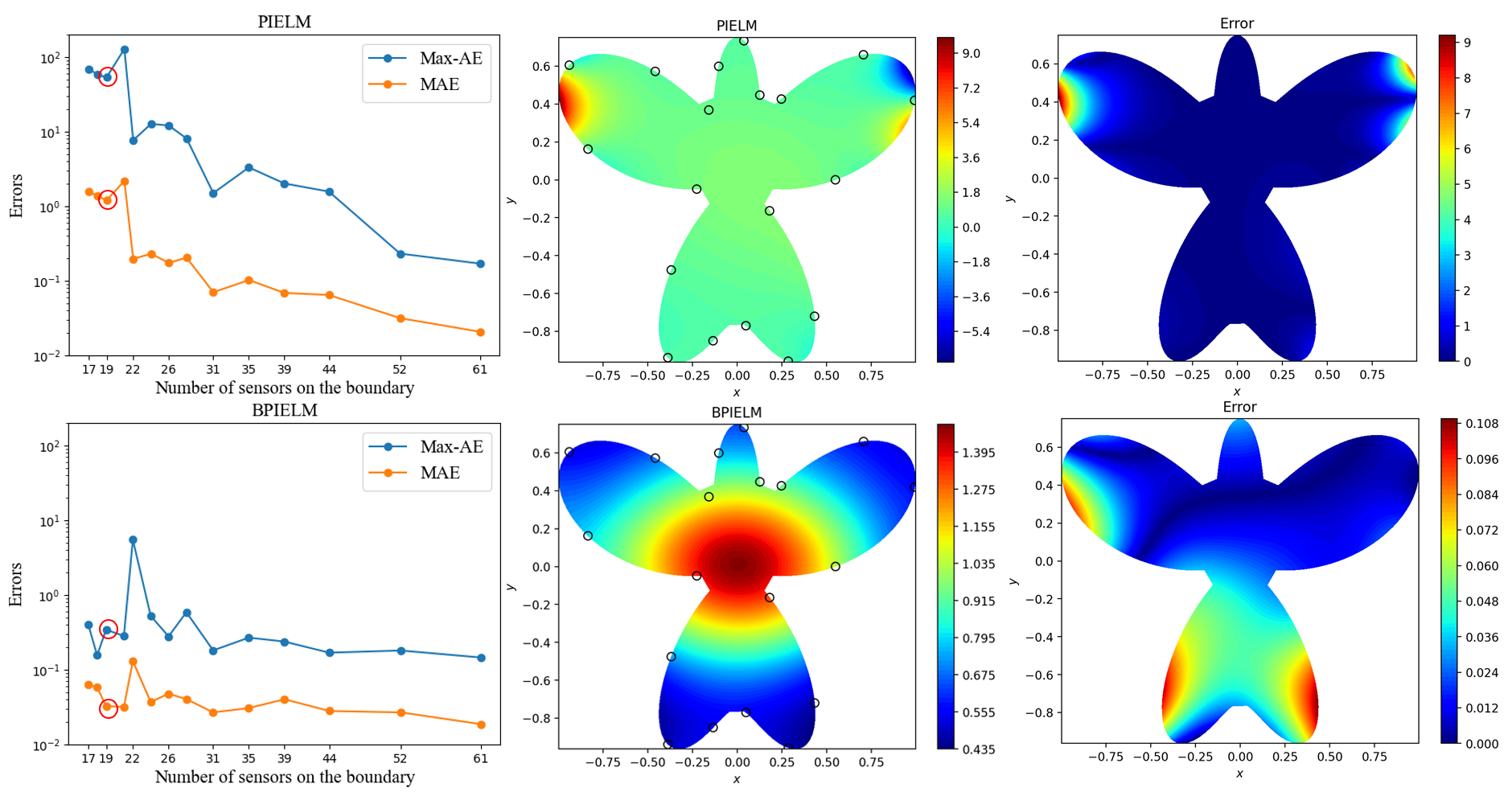}
	\caption{Poisson equation: The first column represents the Max-AE and the MAE under different number of sensors on the boundary, where $\epsilon_{b} \sim \mathcal{N}\left(0,0.01^{2}\right)$, $N=100$ and $N_{f}=200$ are used. The second and third columns represent the predictions at $N_{b}=19$ and its absolute error. The rows (from top to down) represent the results of PIELM and BPIELM. Black circles represent the positions of sensors.}
	\label{fig:Poisson_bc}
\end{figure*}

The effect of the different number of sensors on the boundary for PIELM and BPIELM is illustrated in Fig.\ref{fig:Poisson_bc}. In this test, the number of neurons is $N=100$, the number of training points is $N_{f}=200$, and the noise scale is $\epsilon_{b} \sim \mathcal{N}\left(0,0.01^{2}\right)$. In the first column, we compare the maximum absolute error on the whole domain (Max-AE) and the MAE of PIELM and BPIELM. BPIELM achieves much better MAEs than PIELM with a smaller number of sensors. Besides, BPIELM achieves smaller Max-AEs than PIELM at the same number of sensors. This is because the pseudo-inverse operation is used in PIELM so as to be prone to overfit noisy data (sensor measurements on the boundary), causing the abnormal predictions at some points, but BPIELM alleviates overfitting to decrease the Max-AE. Especially, the results of PIELM and BPIELM with 19 sensors are shown in the last two columns. The errors of PIELM on the two wings of the butterfly tend to exceed $9$, but BPIELM reduces the error to be less than $0.1$. In summary, the above results show that the number of sensors has a stronger effect for PIELM on the predictive accuracy than BPIELM.

\begin{figure*}[!htbp]
	\centering
	\includegraphics[width=1\linewidth]{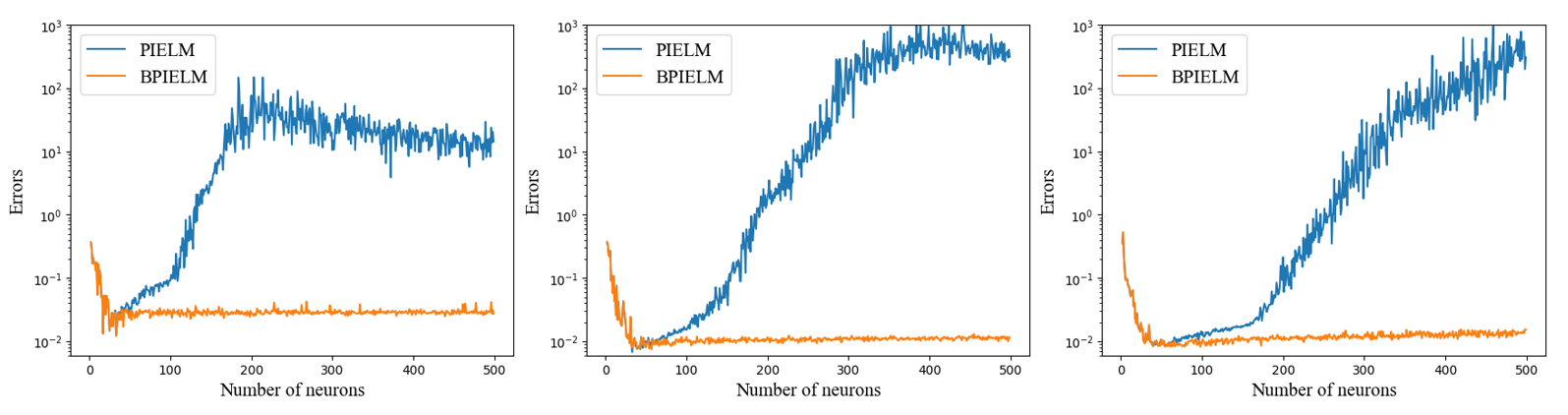}
	\caption{Poisson equation: The columns (from left to right) represent the MAE comparisons between BPIELM and PIELM at $N_{bc}=21$, $N_{bc}=42$ and $N_{bc}=63$, where $\epsilon_{b} \sim \mathcal{N}\left(0,0.01^{2}\right)$ and $N_{f}=400$ are used.}
	\label{fig:Poisson_neuron}
\end{figure*}

PIELM is unstable for the neuron number and is easily affected by noisy data. To this end, we compare the effect of the different neuron number for a fixed number of training points in PIELM and BPIELM on the predictive accuracy. The results are shown in Fig.\ref{fig:Poisson_neuron}. Under noise scenarios, PIELM hardly achieves the best MAE when the number of neurons is equal to or close to the number of constraints, i.e., $N=N_{f}+N_{b}$. The solution accuracy of PIELM is sensitive to the number of neurons. Under the small neuron number, MAEs between PIELM and BPIELM are close. It means that a shallow NN architecture both for PIELM and BPIELM can not learn enough physics information to depict the whole solution. As the neuron number increases, MAEs of PIELM increase after reaching the minimum value in a certain range, while MAEs of BPIELM converge to a small value and fluctuate in a small range. It indicates that an unusually large neuron number in PIELM causes overfitting noisy data, but BPIELM alleviates the overfitting. Especially for $N>N^{*}$, BPIELM gives much better predictions for the solution than PIELM.


Table \ref{tab:Poisson_PINN} is a further comparison of PINN, PIELM and BPIELM under three noise scales, i.e., $\epsilon_{b} \sim \mathcal{N}\left(0,0.01^{2}\right)$, $\epsilon_{b} \sim \mathcal{N}\left(0,0.05^{2}\right)$ and $\epsilon_{b} \sim \mathcal{N}\left(0,0.1^{2}\right)$. In terms of the computational cost (the network training time), PINN takes around 140 seconds to train with Adam \cite{kingma2014adam}. In contrast, the training times of PIELM and BPIELM are on the order of a second and on the order of two seconds, respectively. For the accuracy of the solution, BPIELM provides more accurate predictions than PIELM. The MAE of BPIELM is better than PINN, while PINN is superior to BPIELM in terms of the Max-AE. In a word, BPIELM is not only much more accurate than PIELM but also considerably cheaper than PINN in terms of the computational cost.

\begin{table}[!tbp]
\caption{Poisson equation: The comparisons of PINN, PIELM and BPIELM under three noise scales, i.e., $\epsilon_{b} \sim \mathcal{N}\left(0,0.01^{2}\right)$, $\epsilon_{b} \sim \mathcal{N}\left(0,0.05^{2}\right)$ and $\epsilon_{b} \sim \mathcal{N}\left(0,0.1^{2}\right)$. The number of training points is $N_{f}=400$, $N_{b}=19$ and the number of neurons is $N=100$. PINN has four hidden layers with 100 neurons in each layer and the epoch is 3,000.}
\scalebox{0.92}{
\begin{tabular}{cccccccccccc}
\hline
\multirow{2}{*}{\begin{tabular}[c]{@{}c@{}}Noise \\ scale\end{tabular}} & \multicolumn{3}{c}{0.01}          &  & \multicolumn{3}{c}{0.05}               &  & \multicolumn{3}{c}{0.1}                \\ \cline{2-4} \cline{6-8} \cline{10-12}   & Max-AE & MAE   & Time/s &  & Max-AE & MAE   & Time/s &  & Max-AE & MAE   & Time/s \\ \hline
PINN & 0.165  & 0.043 & 142.30                &  & 0.251  & 0.055 & 141.02                &  & 0.327  & 0.072 & 143.32   \\PIELM  & 0.447  & 0.029 & 1.21                  &  & 0.998  & 0.065 & 1.22                  &  & 1.41   & 0.092 & 1.22\\
BPIELM   & 0.232  & 0.019 & 2.05                  &  & 0.516  & 0.047 & 2.67                  &  & 0.759  & 0.067 & 2.65    \\ \hline
\end{tabular}
}
\label{tab:Poisson_PINN}
\end{table}

\subsection{Advection equation}
The advection equation in one and two spatial dimensions (plus time) is given by 
\begin{equation}
\begin{array}{l}
\frac{\partial u}{\partial t}-c \frac{\partial u}{\partial x}=0, (x, t) \in \Omega, \\
u\left(x_{1}, t\right)=u\left(x_{2}, t\right), u(x, 0)=h(x),
\end{array}
\end{equation}
where $u(x,t)$ is the solution, the constant $c=-2$ is the wave speed, $x_{1}=0$ as well as $x_{2}=1$ are the boundary points, and $\Omega=\{(x, t) \mid x \in [0, 1], t \in [0,1]\}$. The exact solution is considered as
\begin{equation}
u(x, t)=2 \operatorname{sech}\left[3\left(-\frac{1}{2}+\xi\right)\right],
\end{equation}
where 
\begin{equation}
\xi=\bmod \left(x-x_{0}+c t+\frac{L}{2}, L\right), \quad L=x_{2}-x_{1}, \quad x_{0}=0.5.
\end{equation} 
Assuming that the boundary conditions ($u(x_{1},t)$, $u(x_{2},t)$ and $h(x)$) are unknown, we only have $N_{b}$ sensors on the boundary, which are equidistantly distributed on $\partial\Omega$, and consider the Gaussian noise in the measurements.

\begin{figure}[!t]
	\centering
	\includegraphics[width=1\linewidth]{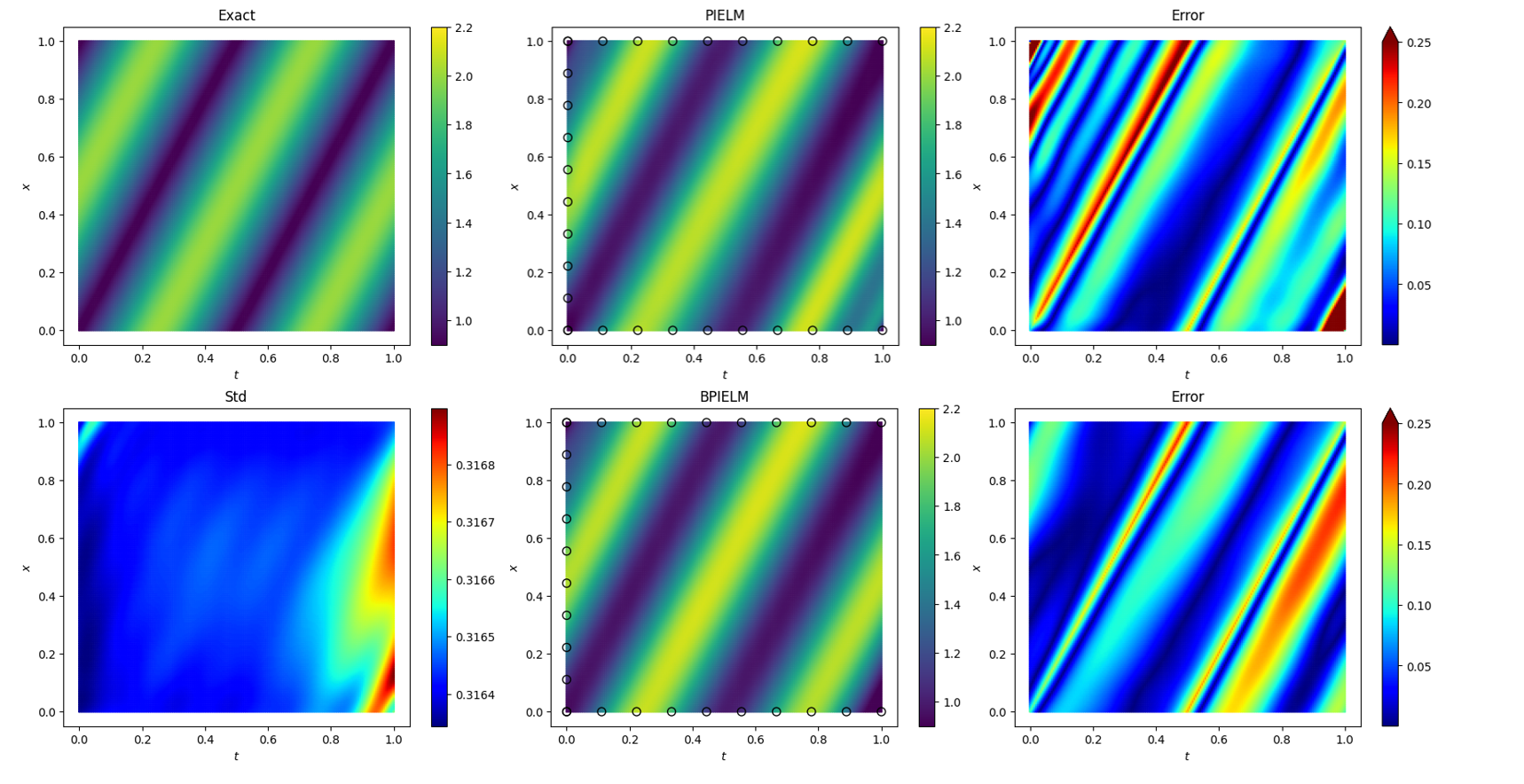}
	\caption{Advection equation: The first row represents the exact solution, the PIELM solution and its absolute error, respectively. The second row represents standard deviations for $u$ using BPIELM, the BPIELM solution and its absolute error. The data noise scale on the boundary is $\epsilon_{b} \sim \mathcal{N}\left(0,0.1^{2}\right)$, the number of neurons is $N=150$ and the number of training points is $N_{f}=400$ and $N_{b}=28$. Black circles represent the positions of sensors.}
	\label{fig:Advection_pred}
\end{figure}

\begin{figure*}[!htbp]
	\centering
	\includegraphics[width=0.9\linewidth]{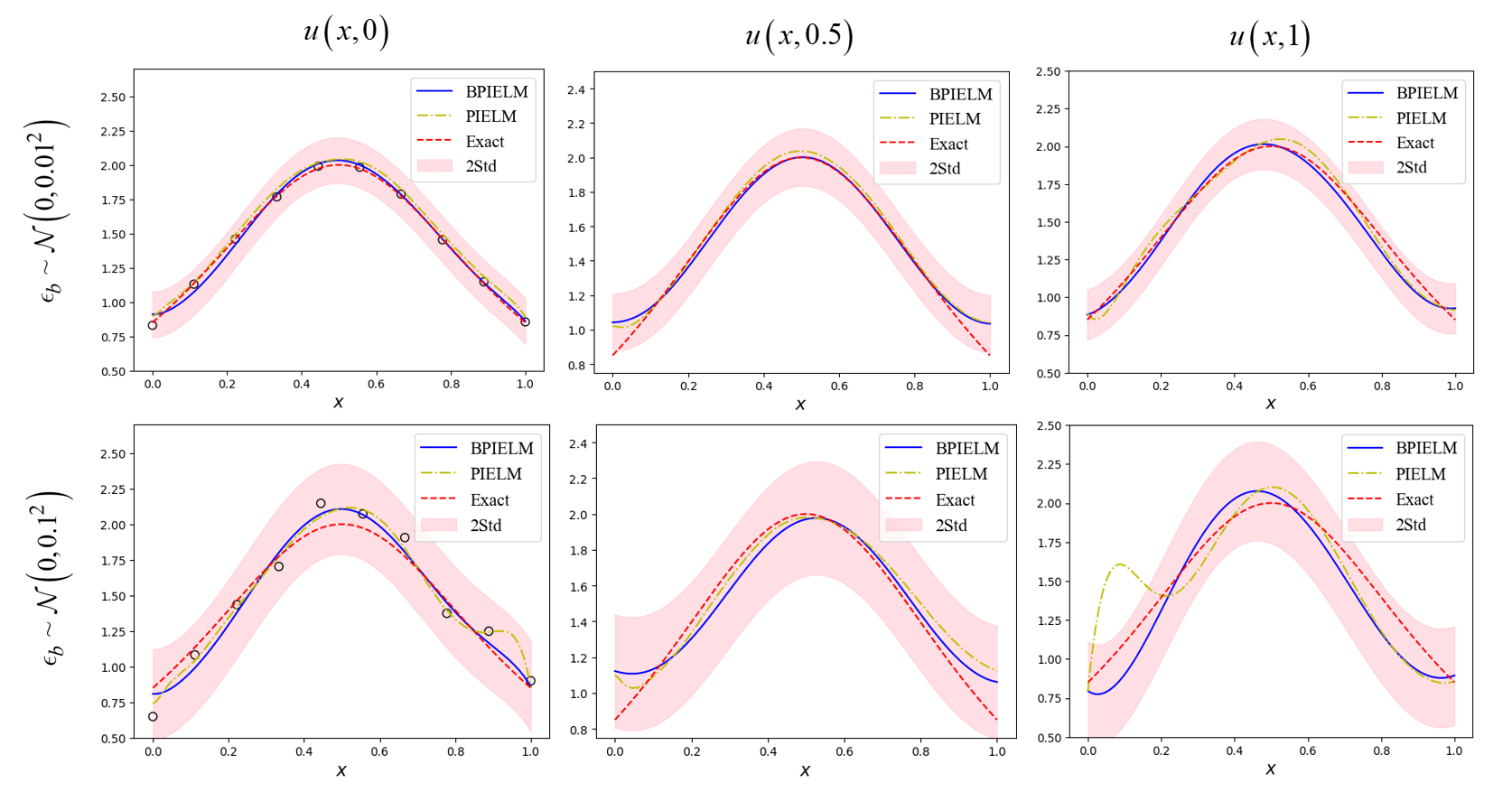}
	\caption{Advection equation: The columns (from left to right) represent comparisons between BPIELM and PIELM at $t=0$, $t=0.1$ and $t = 1$, where $N=150$, $N_{f}=400$ and $N_{b}=28$ are used. The rows (from top to down) represent two noise scales on the boundary, $\epsilon_{b} \sim \mathcal{N}\left(0,0.01^{2}\right)$ and $\epsilon_{b} \sim \mathcal{N}\left(0,0.1^{2}\right)$. }
	\label{fig:Advection_t}
\end{figure*}

To employ BPIELM for solving the advection equation, we encode the governing equations and sensor measurements on the boundary into the linear system, and then use the Bayesian method to estimate the output layer weights. In this simulation, the number of neurons is $N=150$ and the number of training points is $N_{f}=400$ and $N_{b}=28$. The noise scale on the boundary is $\epsilon_{b} \sim \mathcal{N}\left(0,0.1^{2}\right)$. The results of PIELM and BPIELM are illustrated in Fig.\ref{fig:Advection_pred}, where the data noise scale on the boundary is $\epsilon_{b} \sim \mathcal{N}\left(0,0.1^{2}\right)$. Under the large noise scale, BPIELM provides better predictions for the solution than PIELM and gives the uncertainty quantification for the solution. Concretely, MAEs of PIELM and BPIELM are $0.070$ and $0.066$, respectively. For BPIELM, the area with large errors corresponds to relatively large standard deviations. Furthermore, according to three temporal snapshots $t=0,0.5,1$, Fig.\ref{fig:Advection_t} provides the comparisons between PIELM and BPIELM under two different noise scales, i.e., $\epsilon_{b} \sim \mathcal{N}\left(0,0.01^{2}\right)$ and $\epsilon_{b} \sim \mathcal{N}\left(0,0.1^{2}\right)$. The errors of BPIELM are mostly bounded by two standard deviations, and the errors between predictions and the exact solution increase with the increasing noise scale. Under the small noise scale, BPIELM and PIELM provide predictions close to the exact solution. When the noise scale is large, BPIELM gives much better predictions than PIELM, especially for the area near the boundary.

\begin{figure}[!t]
	\centering
	\includegraphics[width=1\linewidth]{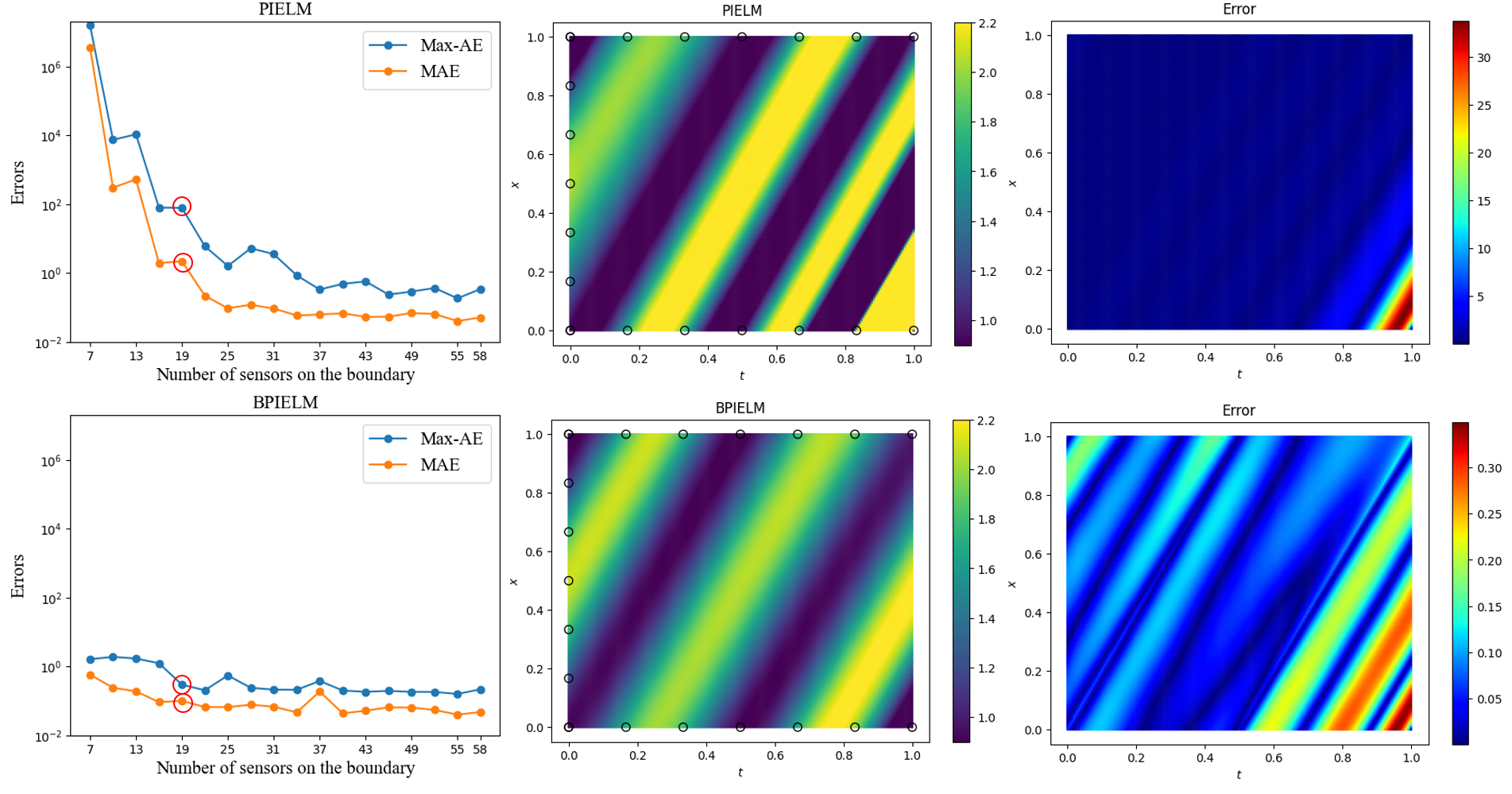}
	\caption{Advection equation: The first column represents the maximum error and MAE under different number of sensors on the boundary, where $\epsilon_{b} \sim \mathcal{N}\left(0,0.1^{2}\right)$, $N=150$ and $N_{f}=400$ are used. The second and third columns represent the prediction at $N_{b}=19$ and its absolute error. The rows (from top to down) represent the results of PIELM and BPIELM. Black circles represent the positions of sensors.}
	\label{fig:Advection_bc}
\end{figure}

Fig.\ref{fig:Advection_bc} illustrates the effect of the different number of sensors on the boundary for PIELM and BPIELM. In this test, the number of neurons is $N=150$, the number of training points is $N_{f}=400$, and the noise scale is $\epsilon_{b} \sim \mathcal{N}\left(0,0.1^{2}\right)$. In the first column, we compare Max-AEs and MAEs of PIELM and BPIELM. Max-AEs and MAEs of PIELM are much worse than BPIELM at the same number of sensors, especially for the scenario with the small number of sensors. In particular, the last two columns show the predictions of PIELM and BPIELM with 19 sensors. The Max-AE of PIELM is extremely large, but BPIELM reduces the Max-AE to be less than $0.4$. In a word, BPIELM outperforms PIELM at the same number of sensors in terms of Max-AEs and MAEs.

\begin{figure*}[!tbp]
	\centering
	\includegraphics[width=1\linewidth]{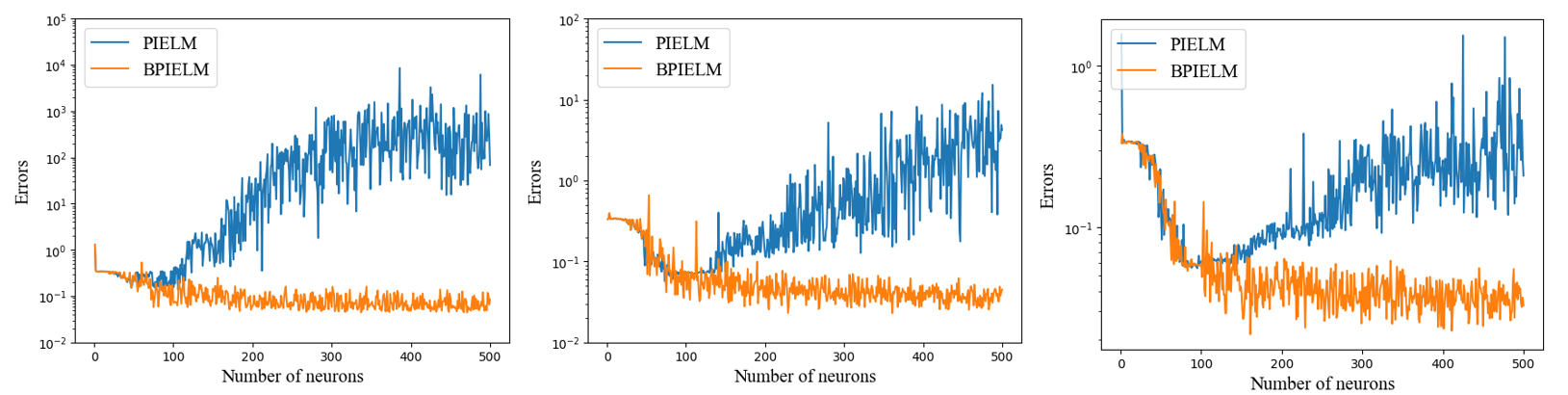}
	\caption{Advection equation: The columns (from left to right) represent comparisons of MAEs between BPIELM and PIELM at $N_{bc}=16$, $N_{bc}=28$ and $N_{bc}=37$, where $\epsilon_{b} \sim \mathcal{N}\left(0,0.1^{2}\right)$ and $N_{f}=400$ are used.}
	\label{fig:Advection_neuron}
\end{figure*}

To further study the impact of the different neuron number on PIELM and BPIELM, we choose three scenarios with $N_{bc}=16,28,37$. The results are shown in Fig.\ref{fig:Advection_neuron}. Under the noise scenario, PIELM hardly achieves the best MAE when the number of neurons is equal to or close to the number of constraints, i.e., $N=N_{f}+N_{b}$. Obviously, the solution accuracy of PIELM is sensitive to the number of neurons, while BPIELM gives much better predictions when the neuron number is larger enough. Moreover, as shown in the figure (from left to right), when the number of noisy measurements increases, MAEs of PIELM and BPIELM decreases and the fluctuation amplitude of the curve increases. However, the fluctuation amplitude of the curve using BPIELM is much smaller than that using PIELM, which shows that BPIELM alleviates overfitting noisy data.

Table \ref{tab:Advection_PINN} presents further comparisons of PINN, PIELM and BPIELM with regard to the accuracy and computational cost under three noise scales. In terms of MAEs, BPIELM is much more accurate than PIELM and PINN, while Max-AEs of BPIELM are close to PINN. For PINN, PIELM and BPIELM, Max-AEs and MAEs increase as the noise increase. Besides, PIELM and BPIELM are dramatically faster to train PINN (by two orders of magnitude).

\begin{table}[htbp]
\caption{Advection equation: The comparisons of PINN, PIELM and BPIELM under three noise scales. The number of training points is $N_{f}=400$, $N_{b}=28$ and the number of neurons is $N=150$. PINN has four hidden layers with 100 neurons in each layer and the epoch is 4,000.}
\scalebox{0.92}{
\begin{tabular}{cccccccccccc}
\hline
\multirow{2}{*}{\begin{tabular}[c]{@{}c@{}}Noise \\ scale\end{tabular}} & \multicolumn{3}{c}{0.01}               &  & \multicolumn{3}{c}{0.05}               &  & \multicolumn{3}{c}{0.1}                \\ \cline{2-4} \cline{6-8} \cline{10-12}  & Max-AE & MAE   & Time/s &  & Max-AE & MAE   & Time/s &  &Max-AE & MAE &Time/s\\\hline
PINN    & 0.162  & 0.033 & 153                   &  & 0.183  & 0.055 & 151   &  & 0.223  & 0.082 & 153  \\PIELM   & 0.442  & 0.037 & 0.98                  &  & 0.541  & 0.047 & 1.01                  &  & 0.670  & 0.070 & 1.22   \\BPIELM    & 0.170  & 0.027 & 1.12                  &  & 0.209  & 0.039 & 1.18                  &  & 0.224  & 0.066 & 1.30                  \\ \hline
\end{tabular}
}
\label{tab:Advection_PINN}
\end{table}


\subsection{Diffusion equation}
The diffusion equation in one and two spatial dimensions (plus time) is given by 
\begin{equation}
\begin{array}{l}
\frac{\partial u}{\partial t}-v \frac{\partial^{2} u}{\partial x^{2}}=f(x, t), (x, t) \in \Omega,\\
u\left(x_{1}, t\right)=b_{1}(t), u\left(x_{2}, t\right)=b_{2}(t), \\
u(x, 0)=h(x),
\end{array}
\end{equation}
where $f(x,t)$ is a source term, the constant $v=0.01$ is the diffusion coefficient, and $\Omega=\{(x, t) \mid x \in [0, 1], t \in [0,2]\}$. The expression for $f(x,t)$ is constructed by choosing the following the exact solution,
\begin{equation}
u(x, t)=\left[2 \cos \left(\pi x+\frac{\pi}{5}\right)+\frac{3}{2} \cos \left(2 \pi x-\frac{3 \pi}{5}\right)\right]\left[2 \cos \left(\pi t+\frac{\pi}{5}\right)+\frac{3}{2} \cos \left(2 \pi t-\frac{3 \pi}{5}\right)\right].
\end{equation}
Assuming that the boundary conditions ($b_{1}(t)$, $b_{2}(t)$ and $h(x)$) are unknown, we only have $N_{b}$ sensors on the boundary, which are equidistantly distributed on $\partial\Omega$, and consider the Gaussian noise in the measurements. 

\begin{figure*}[!t]
	\centering
	\includegraphics[width=1\linewidth]{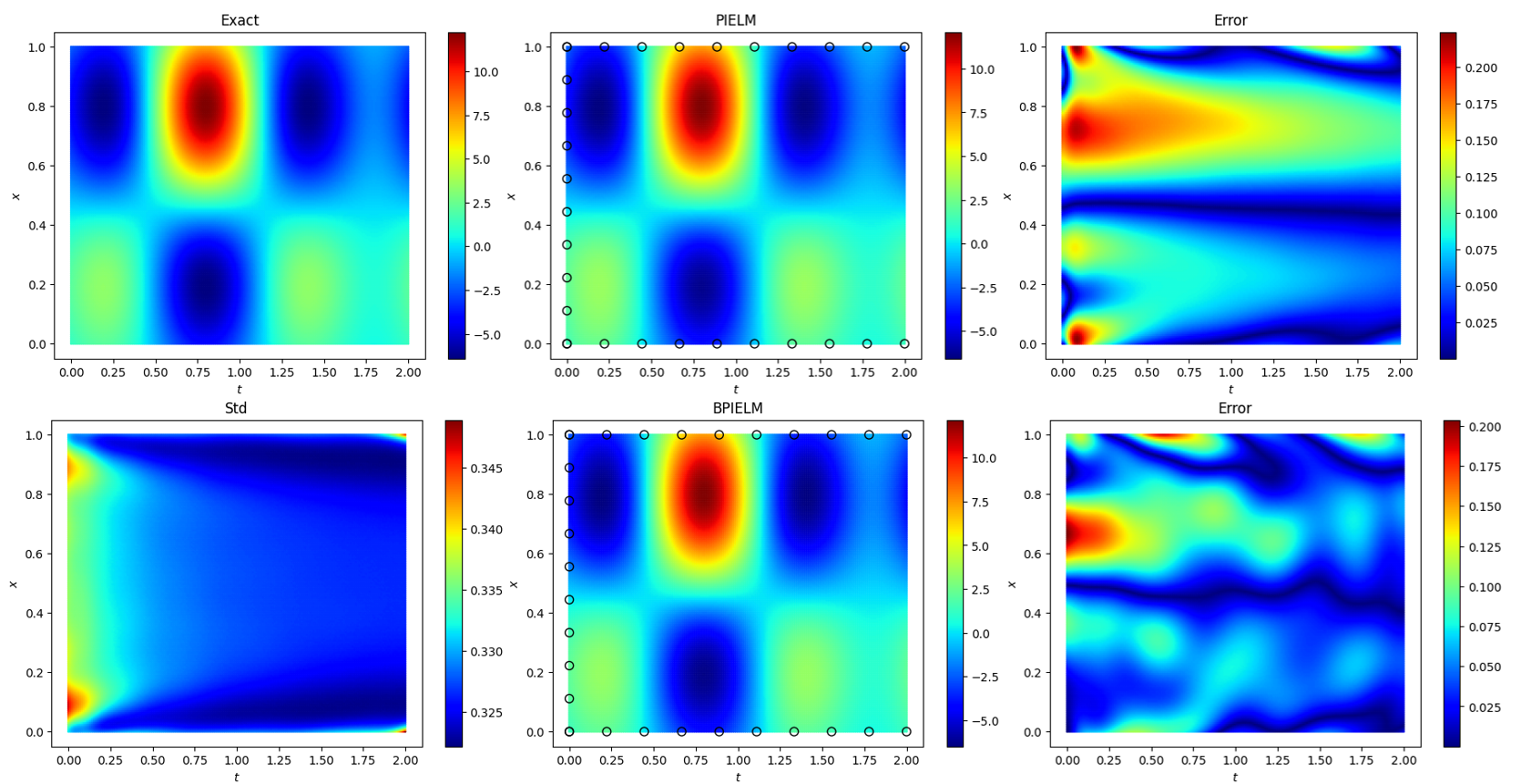}
	\caption{Diffusion equation: The first row represents the exact solution, the PIELM solution and its absolute error, respectively. The second row represents the standard deviations for u using BPIELM, the BPIELM solution and its absolute error, respectively. The data noise scales on the boundary is $\epsilon_{b} \sim \mathcal{N}\left(0,0.1^{2}\right)$, the number of neurons is $N=180$ and the number of training points is $N_{f}=400$ and $N_{b}=28$. Black circles represent positions of sensors.}
	\label{fig:Diffusion_pred}
\end{figure*}

\begin{figure*}[!htbp]
	\centering
	\includegraphics[width=0.9\linewidth]{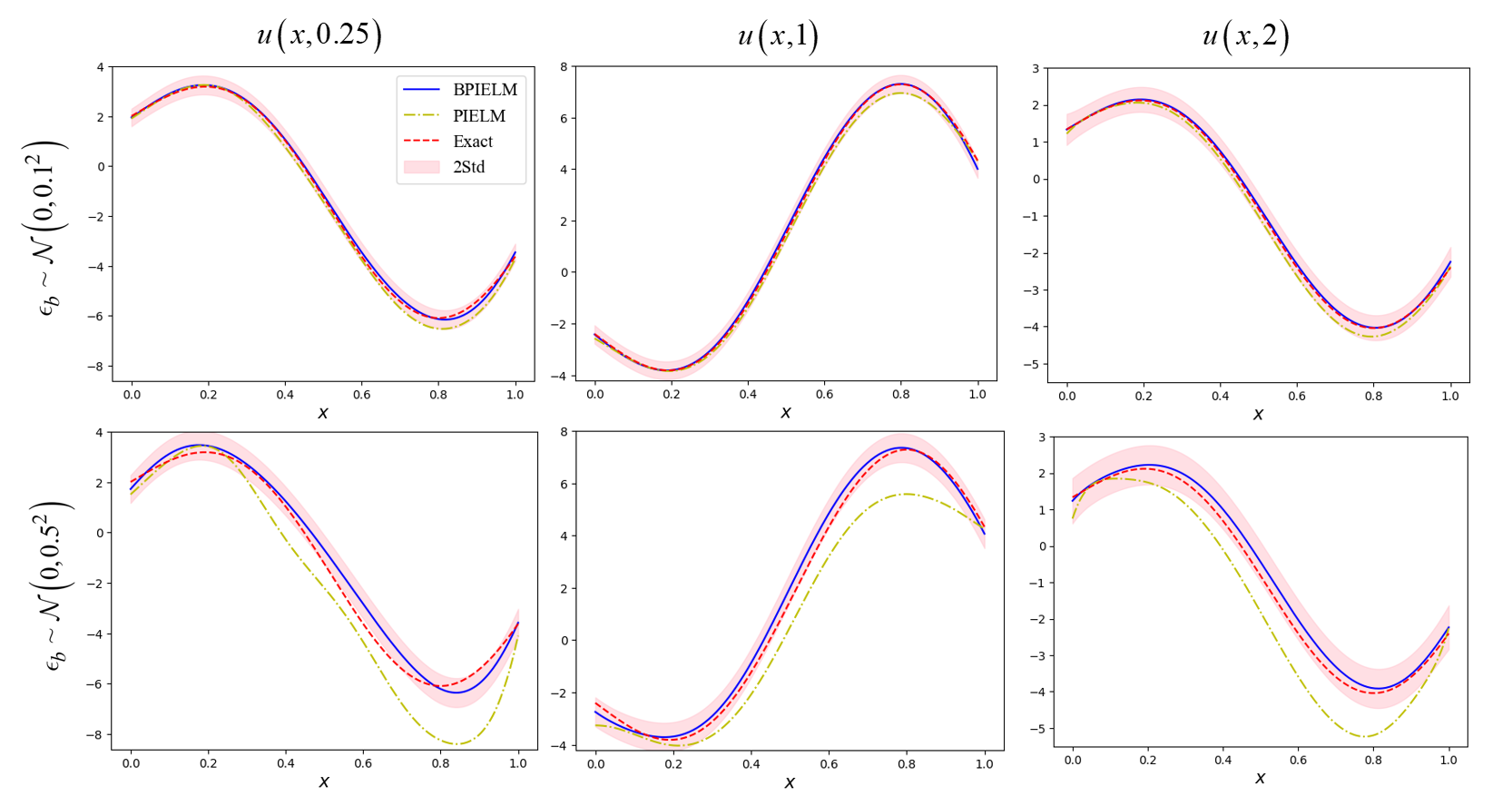}
	\caption{Diffusion equation: The columns (from left to right) represent the comparisons between BPIELM and PIELM at $t=-0.25$, $t=0.1$ and $t = 2$, where $N=180$, $N_{f}=400$ and $N_{b}=30$ are used. The rows (from top to down) represent two noise scales on the boundary, $\epsilon_{b} \sim \mathcal{N}\left(0,0.1^{2}\right)$ and $\epsilon_{b} \sim \mathcal{N}\left(0,0.5^{2}\right)$.}
	\label{fig:Diffusion_t}
\end{figure*}

We approximate the solution using BPIELM to solve the diffusion equation. Concretely, the governing equations and the sensor measurements on the boundary are encoded into the linear system, and then the Bayesian method is used to estimate the output layer weights. In this test, the number of neurons is $N=180$ and the number of training points is $N_{f}=400$ and $N_{b}=28$. The noise scale on the boundary is $\epsilon_{b} \sim \mathcal{N} \left(0,0.1^{2}\right)$. Fig.\ref{fig:Diffusion_pred} illustrates the results of PIELM and BPIELM, where the data noise scale on the boundary is $\epsilon_{b} \sim \mathcal{N}\left(0,0.1^{2}\right)$. BPIELM makes more accurate predictions for the solution than PIELM and gives the uncertainty quantification for the solution. More concretely, MAEs of PIELM and BPIELM are $0.0787$ and $0.0186$, respectively. The absolute errors are concentrated over the area near $t=0$ but also spread all over the spatial-temporal domain. Similarly, for BPIELM, the area near $t=0$ has a relatively large standard deviation. To compare two different noise scales, i.e., $\epsilon_{b} \sim \mathcal{N}\left(0,0.1^{2}\right)$ and $\epsilon_{b} \sim \mathcal{N}\left(0,0.5^{2}\right)$, Fig.\ref{fig:Diffusion_t} provides the results between PIELM and BPIELM at three temporal snapshots $t=0.25,1,2$. As we can see in the figure, the errors of BPIELM are mostly bounded by two standard deviations, and the errors between predictions and the exact solution increases with the increasing noise scale. In particular, when it comes to the large noise scale, the performance of PIELM is much worse than BPIELM.

\begin{figure*}[!t]
	\centering
	\includegraphics[width=0.95\linewidth]{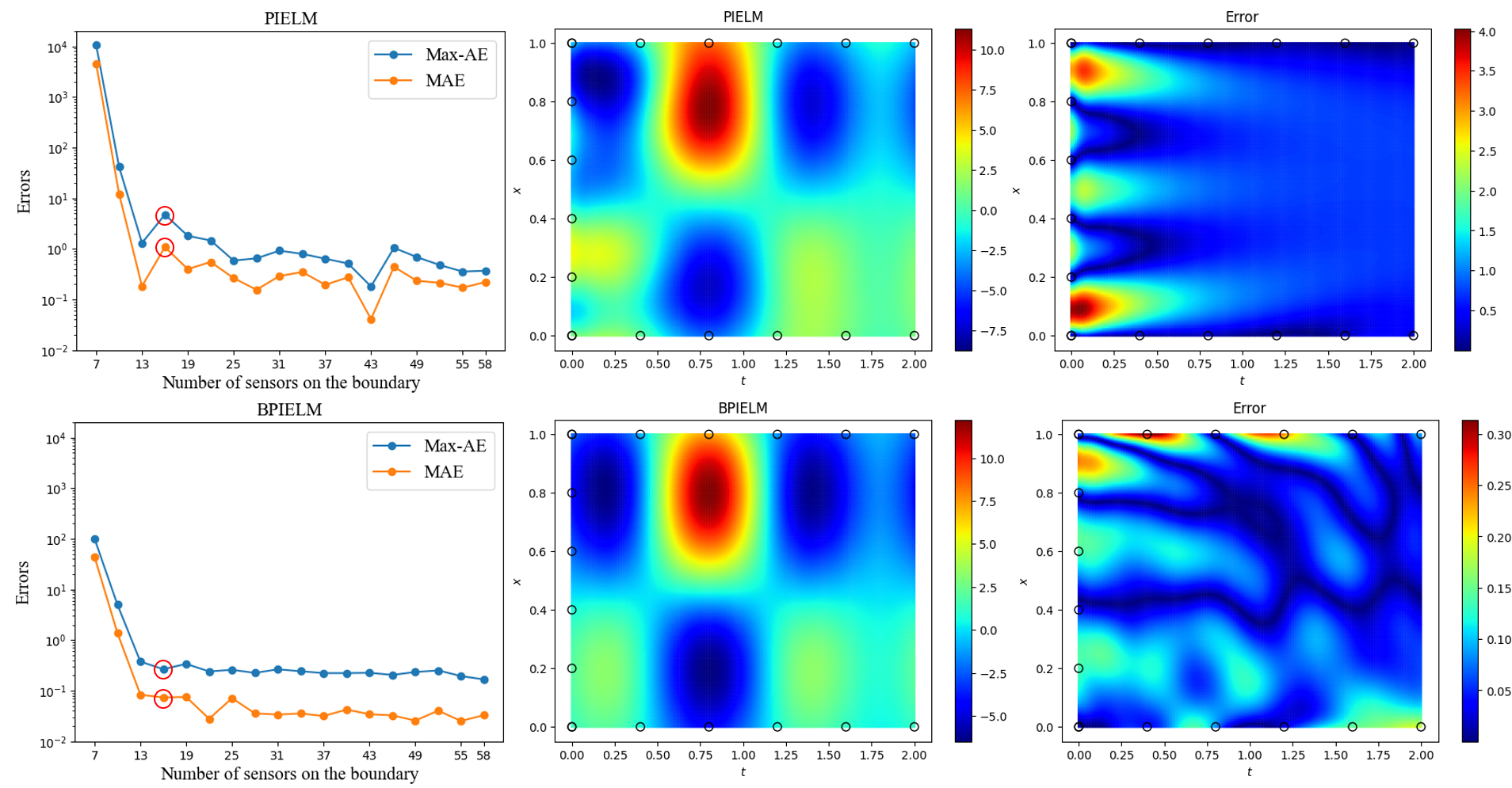}
	\caption{Diffusion equation: The first column represents the maximum error and MAE under different number of measuring points on the boundary, where $\epsilon_{b} \sim \mathcal{N}\left(0,0.1^{2}\right)$, $n=180$ and $N_{f}=400$ are used. The second and third columns represent the predictions at $N_{b}=16$ and its absolute error. The rows (from top to down) represent the results of PIELM and BPIELM. Black circles represent the positions of sensors.}
	\label{fig:Diffusion_bc}
\end{figure*}

To study the effect of the different number of sensors on the boundary for PIELM and BPIELM, we compare the results in Fig.\ref{fig:Diffusion_bc}. In this test, the number of neurons is $N=180$, the number of training points is $N_{f}=400$, and the noise scale is $\epsilon_{b} \sim \mathcal{N}\left(0,0.1^{2}\right)$. As the first column shows, BPIELM outperforms PIELM in terms of Max-AEs and MAEs under the same number of sensors. Max-AEs and MAEs of PIELM and BPIELM are both worse when the number of sensors is extremely small, while MAEs of BPIELM tend to be less than $10^{-1}$ when the number of sensors is more than 13. In particular, the results of PIELM and BPIELM with 16 sensors are shown in the last two columns, where BPIELM provides much more predictions than PIELM.

\begin{figure*}[!t]
	\centering
	\includegraphics[width=1\linewidth]{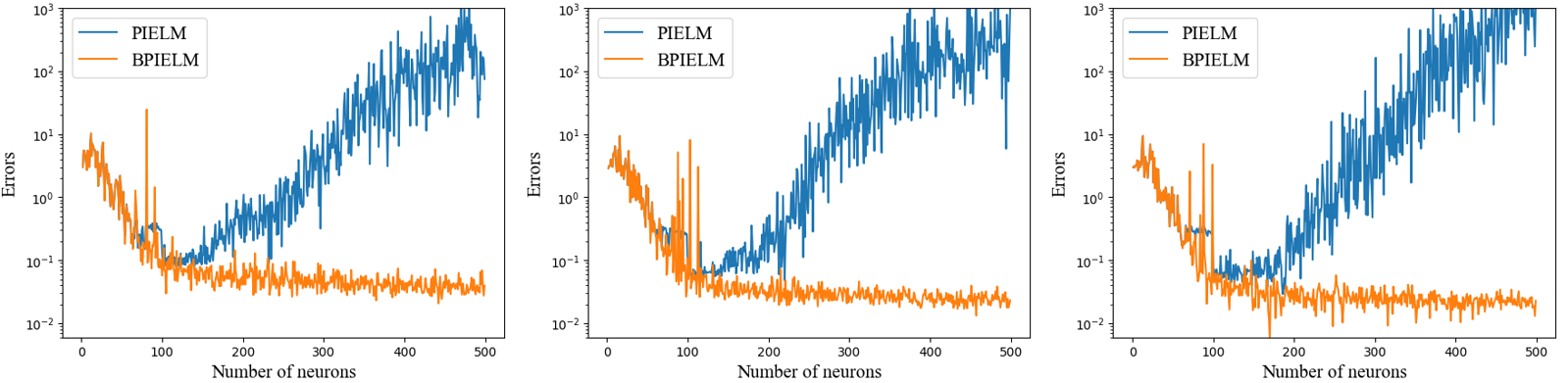}
	\caption{Diffusion equation: The columns (from left to right) represent the MAE comparisons between BPIELM and PIELM at $N_{bc}=42$, $N_{bc}=62$ and $N_{bc}=92$, where $\epsilon_{b} \sim \mathcal{N}\left(0,0.1^{2}\right)$ and $N_{f}=400$ are used.}
	\label{fig:Diffusion_neuron}
\end{figure*}

Fig.\ref{fig:Diffusion_neuron} illustrates the effect of the different neuron number for a fixed number of training points in PIELM and BPIELM on the predictive accuracy. As the figure shows, the MAE curve of PIELM is growing up when $N>200$, while the MAE curve of BPIELM converges to a small value and fluctuates in a small range when the neuron number is large enough. It indicates that the solution accuracy of PIELM is more sensitive to the number of neurons than that of BPIELM.

In Table \ref{tab:Diffusion_PINN}, we provide further comparisons of PINN, PIELM and BPIELM in terms of the accuracy and the computational cost under three noise scales. The computational cost of BPIELM and PIELM is comparable, while BPIELM are more accurate than PIELM. Besides, compared to PINN, the training time of BPIELM and PIELM has been reduced by two orders of magnitude.

\begin{table}[!t]
\caption{Diffusion equation: The comparisons of PINN, PIELM and BPIELM under three noise scales. The number of training points is $N_{f}=400$, $N_{b}=28$ and the number of neurons is $N=180$. PINN has four hidden layers with 100 neurons in each layer and the epoch is 4,000.}
\scalebox{0.92}{
\begin{tabular}{cccccccccccc}
\hline
\multirow{2}{*}{\begin{tabular}[c]{@{}c@{}}Noise \\ scale\end{tabular}} & \multicolumn{3}{c}{0.01}               &  & \multicolumn{3}{c}{0.05}               &  & \multicolumn{3}{c}{0.1}                \\ \cline{2-4} \cline{6-8} \cline{10-12}  & Max-AE & MAE   & Time/s &  & Max-AE & MAE   & Time/s &  & Max-AE & MAE   & Time/s \\ \hline
PINN& 0.082  & 0.019 & 181   &  & 0.138  & 0.032 & 182     &  & 0.182  & 0.033 & 182   \\PIELM  & 0.163  & 0.025 & 0.84 &  & 0.179  & 0.040 & 0.93  &  & 0.226  & 0.079 & 0.91   \\BPIELM  & 0.033  & 0.007 & 1.67 &  & 0.115  & 0.021 & 1.76  &  & 0.118  & 0.019 & 1.68  \\ \hline
\end{tabular}
}
\label{tab:Diffusion_PINN}
\end{table}

\subsection{Inverse problem}

In this subsection, we consider quickly identifying unknown problem parameters in the PDEs by BPIELM. In current research works, PIELM has not been extended to inverse PDE problems, but PIELM can also consider problem parameters as the output layer weights like BPIELM and then uses Moore–Penrose generalized inverse to solve the output layer weights for inverse PDE problems.

\subsubsection{Poisson equation}

As the first inverse problem test, we consider the following linear Poisson equation,
\begin{equation}
\begin{array}{l}
u_{x x}=-\lambda_{1} \cdot \sin (0.7 x)-\lambda_{2} \cdot \cos (1.5 x), \quad x \in \Omega, \\
u\left(x_{1}\right)=b_{1},\quad u\left(x_{2}\right)=b_{2},
\end{array}
\end{equation}
where problem parameters ($\lambda_{1}$ and $\lambda_{2}$) are unknown, $x_{1}$ and $x_{2}$ are boundary points, and $\Omega=\{x \mid x\in[-10,10]\}$. The exact solution is considered in this problem,
\begin{equation}
u=\sin (0.7 x)+\cos (1.5 x)-0.1 x.
\end{equation}
Assuming that the exact solution is unknown, we have $N_{u}$ sensors for $u$, which are equidistantly distributed in $\Omega$. Besides, two sensors are placed at $x=-10$ and $x=10$ to provide the boundary conditions. The Gaussian
noises in the measurements are considered, i.e., $\epsilon_{u}$ and $\epsilon_{b}$.

\begin{figure*}[!t]
	\centering
	\includegraphics[width=1\linewidth]{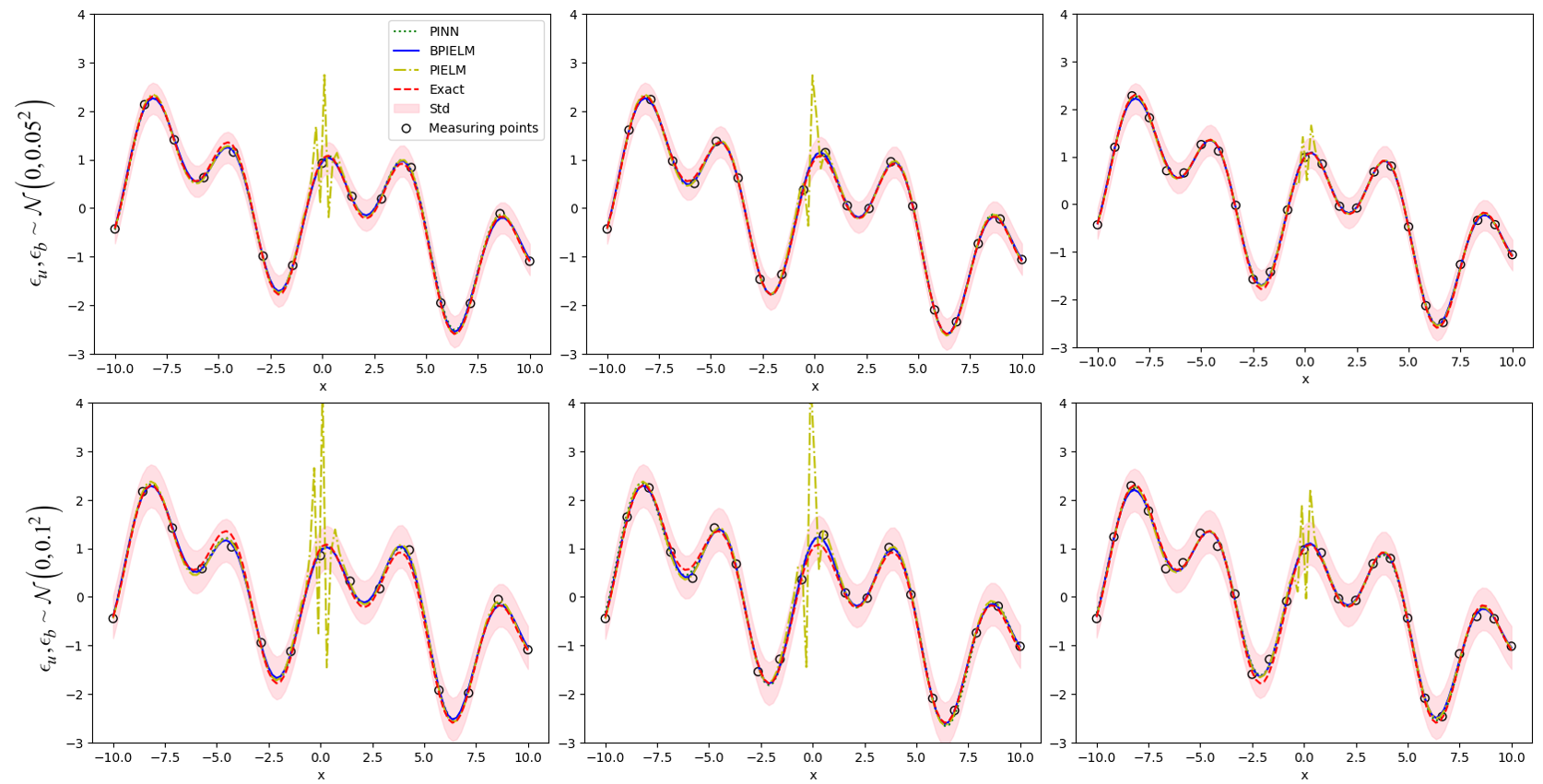}
	\caption{Poisson equation - inverse problem: The comparisons of PINN, PIELM and BPIELM with $N_{u}=13, 18, 23$ under two noise scales. The number of neurons is $N=100$ and the number of training points is $N_{f}=100$ and $N_{b}=2$. Black circles represent the positions of sensors.}
	\label{fig:Poisson_1d_inverse_pred}
\end{figure*}

Let us compare BPIELM with PIELM and PINN for solving the Poisson equation. Fig.\ref{fig:Poisson_1d_inverse_pred} compares the solution using BPIELM, PIELM and PINN under two data noise scales. In the setting of PINN, the NN has three hidden layers with 50 neurons in each layer and the tanh activation function is used. The training epoch with Adam is 4,000. For PIELM and BPIELM, the neuron number is $N=100$ and the number of training points is $N_{f}=100$ and $N_{b}=2$. As shown in the figure, BPIELM gives the uncertainty quantity and provides accurate predictions. The errors of BPIELM are mostly bounded by two standard deviations, which increase as the noise scale increases. Besides, the predictions of PINN are close to BPIELM, which are better than PIELM. Under noise scenarios, PIELM using Moore–Penrose generalized inverse with limited sensor measurements is prone to overfit noisy data, causing worse predictions. The columns from left to right in the figure represent the predictions with $N_{u}=13,18,23$. The predictions with more sensor measurements are better than those with fewer sensor measurements. Besides, the larger noise in the data also causes worse predictions.

Table \ref{tab:Poisson_1d_inverse_vary_noise} provides further comparison with PIELM, BPIELM and PINN for both cases with noise scale, i.e., $\epsilon_{u},\epsilon_{b} \sim \mathcal{N}\left(0,0.05^{2}\right)$ and $\epsilon_{u},\epsilon_{b} \sim \mathcal{N}\left(0,0.1^{2}\right)$. The problem setting here is the same as the Fig.\ref{fig:Poisson_1d_inverse_pred}. The data in the table shows that BPIELM is much more accurate than PIELM and is dramatically faster to train PINN (by two orders of magnitude). BPIELM and PINN give close predictions. Besides, we observe that larger noise leads to worse predictions of the problem parameters and the solution. In a word, the results show the effectiveness of BPIELM in identifying the unknown problem parameters and solving the solution from the scatter noisy data. 

\begin{table}[!t]
\caption{1D Poisson equation - inverse problem: The comparisons of PINN, PIELM and BPIELM under two noise scales. The number of neurons is $N=100$ and the number of training points is $N_{f}=100$, $N_{u}=18$, $N_{b}=2$. The exact solutions for $\lambda_{1}$ and $\lambda_{2}$ are 0.49 and 2.25, respectively.}
\label{tab:Poisson_1d_inverse_vary_noise}
	\centering
	\scalebox{0.92}{
\begin{tabular}{cccccccccccc}
\hline
\multirow{2}{*}{\begin{tabular}[c]{@{}c@{}} Method \end{tabular}} & \multicolumn{5}{c}{$\epsilon_{u},\epsilon_{b} \in 0.05$}              &  & \multicolumn{5}{c}{$\epsilon_{u},\epsilon_{b}\in 0.1$}               \\ \cline{2-6} \cline{8-12}  & $\lambda_{1}$ & $\lambda_{2}$ & Max-AE & MAE   & Time/s &  & $\lambda_{1}$ & $\lambda_{2}$ & Max-AE & MAE   & Time/s \\ \hline
PINN     & 0.491 & 2.307 & 0.073  & 0.027 & 225.30  &  & 0.488 & 2.370 & 0.160  & 0.068 & 240.12  \\ PIELM    & 0.484 & 2.390 & 1.808  & 0.081 & 0.71    &  & 0.477 & 2.280 & 3.619  & 0.168 & 0.77   \\BPIELM  & 0.490 & 2.300 & 0.076  & 0.027 & 2.51    &  & 0.480 & 2.300 & 0.155  & 0.054 & 2.53    \\ \hline
\end{tabular}
}
\end{table}


\begin{figure*}[!htbp]
	\centering
	\includegraphics[width=1\linewidth]{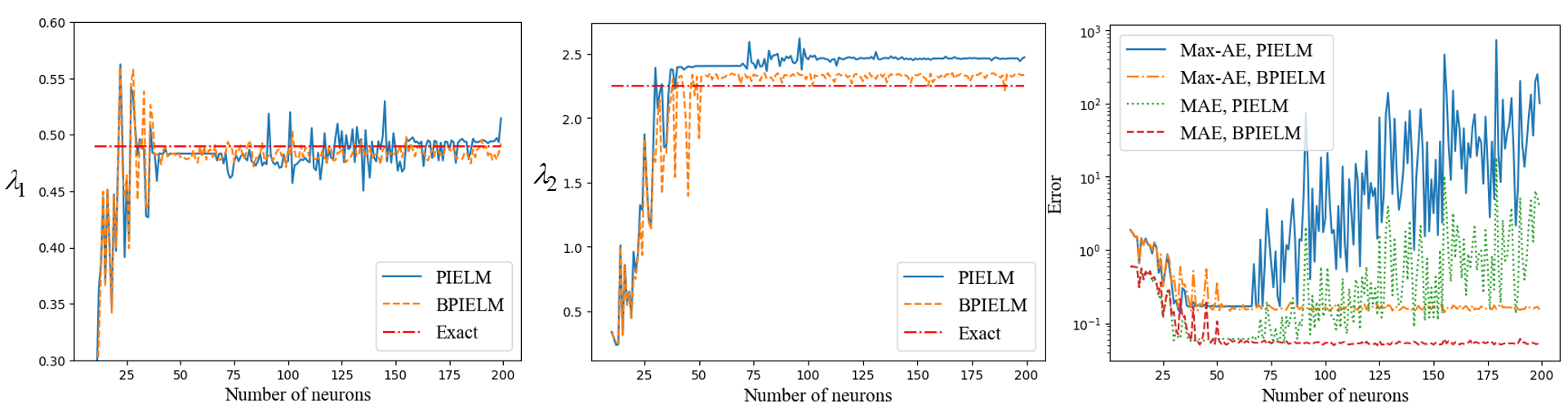}
	\caption{1D Poisson equation - inverse problem: The comparisons of PIELM and BPIELM under the different neuron number. The first two columns represent the predictions for the system parameters, $\lambda_{1}$ and $\lambda_{2}$, where the exact solutions for $\lambda_{1}$ and $\lambda_{2}$ are 0.49 and 2.25. The last column represents the comparisons of Max-AEs and MAEs of PIELM and BPIELM. The number of sensors is $N_{u}=18$, the number of neurons is $N=100$ and the number of training points is $N_{f}=100$ and $N_{b}=2$.}
	\label{fig:Poisson_1d_inverse_vary_neuron}
\end{figure*}

Fig.\ref{fig:Poisson_1d_inverse_vary_neuron} demonstrates the effect of the different neuron number for a fixed number of training points in PIELM and BPIELM on identifying problem parameters and predicting the solution. The data in the first two columns show that BPIELM are more stable for the neuron number, while PIELM is more sensitive to the neuron number. Note that when the neuron number is large enough, the predictive problem parameters of BPIELM converge to exact values and fluctuate in a small range. As the last column shows, Max-AEs as well as MAEs of PIELM and BPIELM are close under the small neuron number and then PIELM achieves the best Max-AEs and MAEs within a certain neuron number, but it is difficult for PIELM to find appropriate neuron number in practice except for the experiments. However, BPIELM alleviates overfitting noisy data and a large enough number of neurons is chosen for BPIELM, which is more appropriate for noise scenarios.


\subsubsection{Helmholtz equation}

In the next test, we consider the following 1D linear Helmholtz equation

\begin{equation}
\begin{array}{l}
u_{x x}+k^{2}u=-\lambda_{1} f_{1}(x)+\lambda_{2} \cdot f_{2}(x)-\lambda_{3}, x \in \Omega, \\
u\left(x_{1}\right)=b_{1},  \quad u\left(x_{2}\right)=b_{2},
\end{array}
\end{equation}
where the problem parameters ($\lambda_{1}$, $\lambda_{2}$ and $\lambda_{3}$) are unknown, the wave number is $k^{2}=10$, $x_{1}$ as well as $x_{2}$ are boundary points, $f_{1}(x)= \sin (2 x)\cos (4 x)$, $f_{2}(x)= \cos (2 x) \sin (4 x)$, and $\Omega=\{x \mid x\in[-2\pi,2\pi]\}$. The exact solution is considered in this problem,
\begin{equation}
u=\sin (2 x)\cos (4 x)+1.
\end{equation}
Assuming that the exact solution is unknown, we have $N_{u}$ sensors for $u$, which are equidistantly distributed in $\Omega$. Besides, two sensors are placed at $x_{1}=-2\pi$ and $x_{2}=2\pi$ to provide the boundary conditions for $u$. The Gaussian noises in the measurements are considered, i.e., $\epsilon_{u}$ and $\epsilon_{b}$.

\begin{figure*}[!t]
	\centering
	\includegraphics[width=1\linewidth]{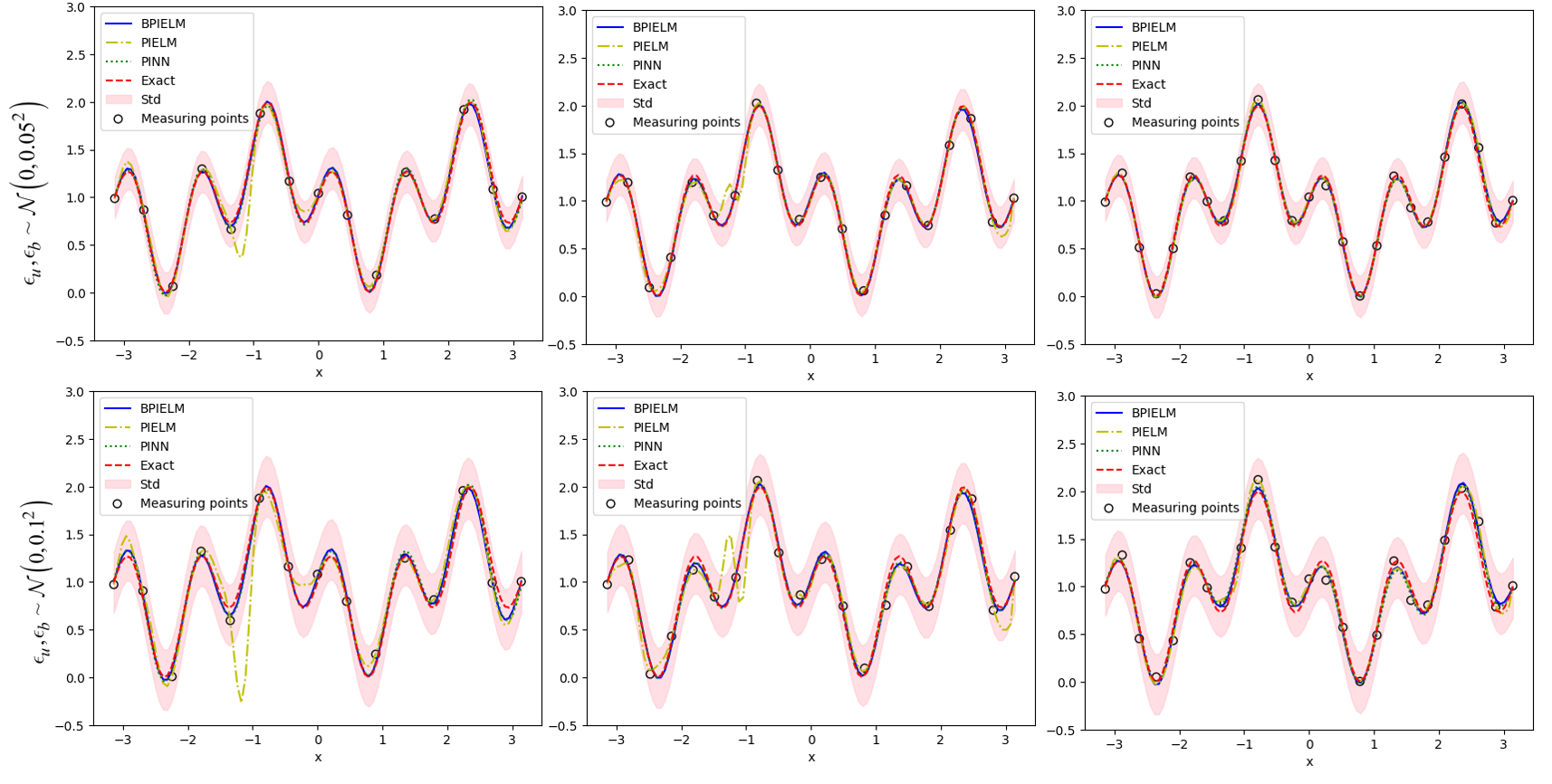}
	\caption{Helmholtz equation - inverse problem: The comparisons of PINN, PIELM and BPIELM with $N_{u}=13, 18, 23$ under two noise scales. The number of training points is $N_{f}=100$ and $N_{b}=2$. Black circles represent the positions of sensors.}
	\label{fig:Helmholtz_1d_inverse_pred}
\end{figure*}

We next compare BPIELM with PIELM and PINN for solving the Helmholtz equation, where more unknown problem parameters are identified. Fig.\ref{fig:Helmholtz_1d_inverse_pred} compares the solution using BPIELM, PIELM and PINN under two data noise scale, .i.e., $\epsilon_{u}, \epsilon_{b} \sim \mathcal{N}\left(0,0.05^{2}\right)$ and $\epsilon_{u}, \epsilon_{b} \sim \mathcal{N}\left(0,0.1^{2}\right)$. In the setting of PINN, we use four hidden layers with 100 neurons in each layer and use 4,000 epochs to train the NN with the tanh activation function. For PIELM and BPIELM, the number of neurons is also set to be $N=100$. As the figure shows, BPIELM over PIELM and PINN has a two-fold benefit: (1) BPIELM gives the uncertainty quantity for the solution and errors are mostly bounded by two standard deviations. (2) BPIELM provides more accurate predictions for the solution than PIELM and BPIELM reduces the computational cost to achieve close or better predictions than PINN (the computational cost is shown in Table \ref{tab:Helmholtz_1d_inverse_vary_noise}). 

Table \ref{tab:Helmholtz_1d_inverse_vary_noise} provides further comparison with PIELM, BPIELM and PINN for both cases with noise scale, i.e., $\epsilon_{u}, \epsilon_{b} \sim \mathcal{N}\left(0,0.05^{2}\right)$ and $\epsilon_{u}, \epsilon_{b} \sim \mathcal{N}\left(0,0.1^{2}\right)$. The problem setting here is the same as the Fig.\ref{fig:Helmholtz_1d_inverse_pred}. The results show the effectiveness of BPIELM in identifying the unknown problem parameters and solving the solution from the scatter noisy data. BPIELM is not only much more accurate than PIELM, but also considerably cheaper than PINN in terms of the computational cost. The training time with the BPIELM is on the order of two seconds. In contrast, it is takes 400 seconds to train PINN to obtain close predictions.

\begin{figure*}[!t]
	\centering
	\includegraphics[width=0.7\linewidth]{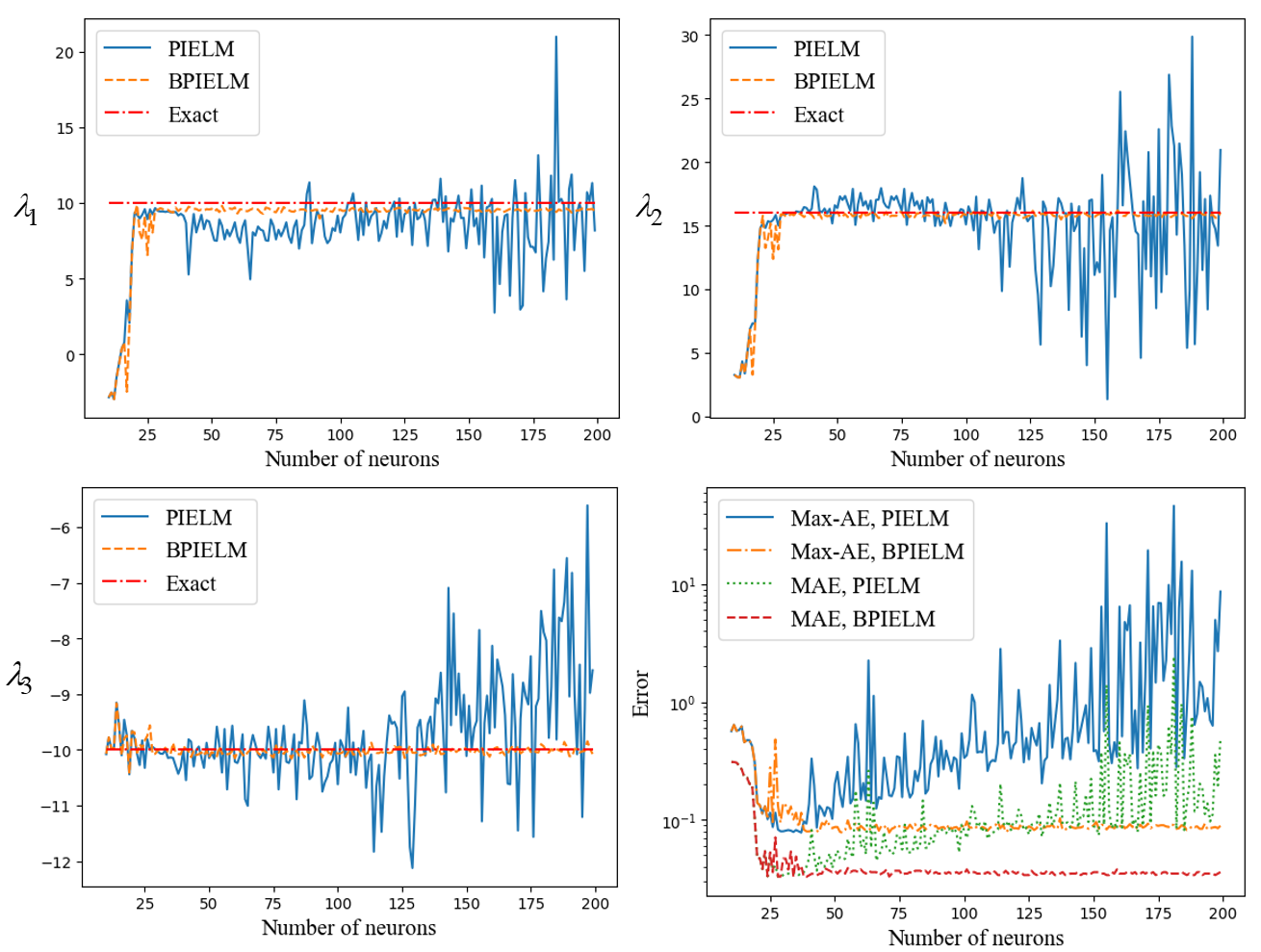}
	\caption{1D Helmholtz equation - inverse problem: The comparisons of PIELM and BPIELM under the different neuron number. The first row represents the predictions for the system parameters, $\lambda_{1}$ and $\lambda_{2}$. The second row (from left to right) represents the prediction for the system parameter $\lambda_{3}$ and comparisons of Max-AEs and MAEs. The number of neurons is $N=100$ and the number of training points is $N_{f}=100$ and $N_{b}=2$. The exact solutions for $\lambda_{1}$, $\lambda_{2}$ and $\lambda_{3}$ are 10, 16, -10, respectively.}
	\label{fig:Helmholtz_1d_inverse_vary_neuron}
\end{figure*}

\begin{table}[!t]
\caption{1D Helmholtz equation - inverse problem: The comparisons of PINN, PIELM and BPIELM under two noise scales. The number of neurons is $N=100$ and the number of training points is $N_{f}=100$, $N_{u}=18$, $N_{b}=2$. The exact solutions for $\lambda_{1}$, $\lambda_{2}$ and $\lambda_{3}$ are 10, 16, -10, respectively.}
\centering
\label{tab:Helmholtz_1d_inverse_vary_noise}
\scalebox{0.81}{
\begin{tabular}{cccccccccccccc}
\hline
\multirow{2}{*}{\begin{tabular}[c]{@{}c@{}} Method \end{tabular}} & \multicolumn{6}{c}{$\epsilon_{u},\epsilon_{b}\in 0.05$}                       &  & \multicolumn{6}{c}{$\epsilon_{u},\epsilon_{b}\in 0.1$}  \\ \cline{2-7} \cline{9-14}   & $\lambda_{1}$ & $\lambda_{2}$ & $\lambda_{3}$ & Max-AE & MAE   & Time/s &  & $\lambda_{1}$ & $\lambda_{2}$ & $\lambda_{3}$ & Max-AE & MAE   & Time/s \\ \hline
PINN    & 9.84  & 15.84 & -9.96 & 0.034  & 0.015 & 399.38                &  & 9.71  & 15.71 & -9.93 & 0.071  & 0.031 & 398.13                \\
PIELM                                                                   & 9.71  & 14.8  & -10   & 0.337  & 0.047 & 0.69                  &  & 9.41  & 13.5  & -10   & 0.683  & 0.096 & 0.81                      \\
BPIELM                                                                  & 9.80  & 16.0  & -10   & 0.051  & 0.018 & 2.59                  &  & 9.60  & 15.9  & -10   & 0.091  & 0.035 &     2.51                  \\ \hline
\end{tabular}
}
\end{table}


Fig.\ref{fig:Helmholtz_1d_inverse_vary_neuron} illustrates the effect of different neuron number for a fixed number of training points in PIELM and BPIELM on identifying problem parameters. The results show that PIELM is dramatically more sensitive to the neuron number than BPIELM. For BPIELM, when the neuron number is large enough, predictive problem parameters converge to exact values and fluctuate in a small range, which indicates the BPIELM avoids overfitting noisy data. For Max-AE and MAE curves, under the small neuron number, MAEs between PIELM and BPIELM are close and worse. It means that a shallow NN architecture both for PIELM and BPIELM can not learn enough physics information to depict the whole solution. As the neuron number increases, MAEs of PIELM increase after reaching the minimum value in a certain range, while the Max-AE and the MAE curves of BPIELM converge to the values and fluctuate in a small range.
In summary, considering the accuracy as well as the sensibility to the neuron number, BPIELM is a better approach than PIELM for noise scenarios.


\section{Conclusion}

\label{sec conclusion}
In this work, we propose the BPIELM framework to quantify the uncertainty arising from noisy data for forward and inverse PDE problems. The physics laws (PDE, boundary conditions, and measurements) are first incorporated into the NN as the cost function. Then a prior probability distribution is introduced in the output layer and the Bayesian method is used to estimate the posterior. To support our claim, we examine BPIELM for forward and inverse PDE problems, including Poisson, advection, diffusion and Helmholtz equations. In those tests under noise scenarios, the results show that BPIELM has a three-fold benefit: (1) BPIELM provides the uncertainty quantification for the solution and more accurate predictions for solutions as well as problem parameters than PIELM. (2) PIELM is prone to be drastically sensitive to the number of hidden neurons, but BPIELM alleviates the sensitivity characteristic and improves the robustness of the NN. (3) BPIELM is considerably cheaper than PINN in terms of the computational cost.

Although BPIELM has exhibited some superior potential for designing fast and accurate scientific machine learning techniques, some things need to be resolved. The high efficiency of PIELM benefits from the linear nature of the final problem, therefore BPIELM, which inherits some characteristics of PIELM, is also limited to studying linear problems. How to further extend and improve this method is an interesting problem, and will be explored in future works.

\section*{CRediT authorship contribution statement}
\textbf{Xu Liu}: Software, Methodology, Formal analysis, Writing - original draft. \textbf{Wen Yao}: Supervision, Funding acquisition, Project administration. \textbf{Wei Peng}: Conceptualization, Supervision. \textbf{Weien Zhou}: Software, Supervision, Data curation, Visualization, Funding acquisition.

\section*{Declaration of competing interest}
The authors declare that they have no known competing financial interests or personal relationships that could have appeared to influence the work reported in this paper.

\section*{Acknowledgements} 
This work was supported by National Natural Science Foundation of China under Grant No.11725211, 52005505, and 62001502.

\bibliography{mybibfile}

\end{document}